%% file: ms.tex
\documentclass[letterpaper, 10 pt, conference]{ieeeconf}  

\IEEEoverridecommandlockouts                              

\usepackage{graphicx}
\usepackage{times} 
\usepackage{amsmath} 
\usepackage{amssymb}  
\usepackage{color}

\usepackage{url}
\usepackage{subcaption}
\usepackage{booktabs}
\usepackage[font={small}]{caption}
\DeclareMathOperator*{\argmin}{argmin}

\DeclareMathOperator*{\cov}{cov}
\title{\LARGE \bf
RaD-VIO: Rangefinder-aided Downward Visual-Inertial Odometry
}

\author{Bo Fu, Kumar Shaurya Shankar, Nathan Michael
\thanks{Authors are with The Robotics Institute, Carnegie Mellon University, Pittsburgh, PA, 15213, USA
        {\tt\small \{bofu, kshaurya, nmichael\}@cmu.edu}}%
}

\begin{document}

\maketitle
\thispagestyle{empty}
\pagestyle{empty}
\input{ieeeconf/abstract}
\input{ieeeconf/introduction}
\input{ieeeconf/vision}
\input{ieeeconf/fusion}
\input{ieeeconf/simulation_experiment}
\input{ieeeconf/conclusion}
\clearpage
\bibliography{ms}
\bibliographystyle{IEEEtran}

\end{document}

%% file: ieeeconf/abstract.tex
\begin{abstract}
State-of-the-art forward facing monocular visual-inertial odometry algorithms are often brittle in practice, especially whilst dealing with initialisation and motion in directions that render the state unobservable. In such cases having a reliable complementary odometry algorithm enables robust and resilient flight. Using the common local planarity assumption, we present a fast, dense, and direct frame-to-frame visual-inertial odometry algorithm for downward facing cameras that minimises a joint cost function involving a homography based photometric cost and an IMU regularisation term. Via extensive evaluation in a variety of scenarios we demonstrate superior performance than existing state-of-the-art downward facing odometry algorithms for Micro Aerial Vehicles (MAVs).
\end{abstract}

%% file: ieeeconf/introduction.tex
\section{Introduction and Related Work}

Recent advances in optimisation based monocular visual-inertial SLAM algorithms for MAVs have made great strides in being accurate and efficient~\cite{delmerico2018benchmark}. However, in practice, these algorithms suffer from three main failure modalities - sensitivity to initialisation, undergoing motion that renders the state unobservable, and, to a lesser extent, inability to handle outliers within the optimisation. The first arises from the need for translation to accurately triangulate feature landmarks and being able to excite all axes of the accelerometer to determine scale. The second is a fundamental limit of the sensor characteristics, robot motion, and the environment, most often caused by motion in the camera direction and an absence of texture information. The third is often an artefact of sliding windows necessitated by the constraints imposed by limited compute on aerial platforms.

We believe that in order to have resilient closed loop flight it is imperative to have complementary sources of odometry. Towards this, we present an algorithm that computes metric velocity without depending on triangulation or feature initialisation, utilises observability in an orthogonal direction to a conventional forward facing camera, and is purely a frame-to-frame method. This enables it to be fast and reliable while still being accurate.

In this paper, we pursue the problem of estimating the linear and angular velocity and orientation of a micro aerial vehicle (MAV) equipped with a downward facing camera, an IMU, and a single beam laser rangefinder which measures the height of the vehicle relative to the ground. 
\begin{figure}[h]
	\centering
	\includegraphics[width=\linewidth]{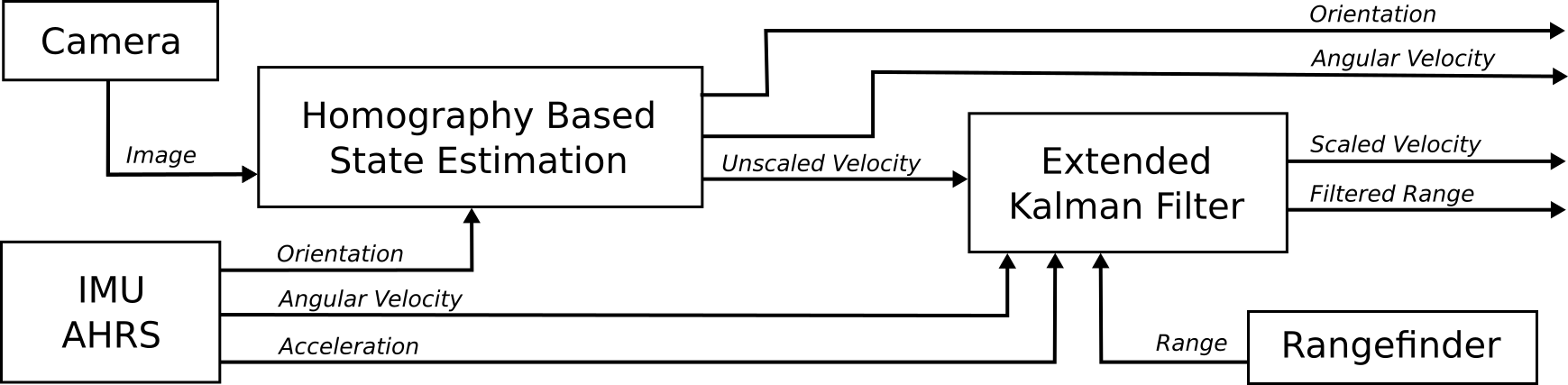}
	\caption{System Diagram. The homography based optimisation takes in images and differential rotation estimates from the Attitude and Heading Reference System (AHRS) on the IMU and outputs an orientation and unscaled velocity, which is then used in an Extended Kalman filter to provide scaled metric velocity.}
\end{figure}

A common strategy for performing visual odometry using downward facing cameras involves exploiting epipolar geometry using loosely-coupled~\cite{weiss2012real,forster2017svo} or tightly-coupled visual-inertial methods~\cite{mourikis2007multi, qin2018vins}. An alternate class of approaches make a planar ground assumption which enables optical flow based velocity estimation where the camera ego motion is compensated using angular rate data obtained from a gyroscope and metric scaling obtained using an altitude sensor~\cite{honegger2013open}. An issue with all such epipolar constraint based methods is that their performance is predicated on there being detectable motion between camera frames. Common failure modes for this class of techniques are situations when the camera is nearly static in hovering conditions or when it moves vertically.

These failure modes can be mitigated by explicitly encoding the planar ground assumption using the homography constraint between successive camera views. Implicit means of utilising this constraint have been presented earlier in appearance based localisation~\cite{steder2008visual} where cameras are localised against a library of previously acquired images. Most relevant to our approach, the authors in \cite{grabe2012board,grabe2013comparison,grabe2015nonlinear} first estimate the optical flow between features in consecutive frames and then explicitly use the homography constraint and the angular velocity and ground plane orientation obtained from an inertial sensor to obtain unscaled velocity. They finally use an extended Kalman filter (EKF) to fuse the data and output metric velocity. We use this work as our baseline.

In this work, instead of using sparse visual features that are highly dependent on textured environments, we utilise a dense, direct method that makes use of all the visual information present in the camera image and couple it with angular constraints provided by an IMU within a least squares optimisation. We then fuse the result of this optimisation with altitude data from a rangefinder to obtain metric velocity.

Contributions of this work include:
\begin{itemize}
	\item A homography based frame-to-frame velocity estimation algorithm, that is accurate and robust in a wide variety of scenes;
	\item An EKF structure to incorporate this with a single beam laser rangefinder signal and estimate IMU bias; and
	\item Extensive evaluation on a wide variety of environments with comparisons with state of the art algorithms.
\end{itemize}

%% file: ieeeconf/vision.tex
\section{Estimation Theory}
In this section we present the homography constraint, our optimisation strategy, and the framework to incorporate the corresponding cost functions.
\subsection{Homography Constraint and Parameterisation}\label{sec:homography_constraint}
When looking at points lying on the same plane, their projected pixel coordinates in two images (\(\mathbf{X}\) and \(\mathbf{X}'\) respectively) taken by a downward camera can be related by
\begin{equation}
	\mathbf X \equiv \mathbf H \mathbf X'
\end{equation}
where
\begin{equation}
	\mathbf H = \mathbf K(\mathbf R + \mathbf t_0/d\cdot \mathbf n^\text{T})\mathbf K^{-1} = \mathbf K(\mathbf R + \mathbf t\cdot \mathbf n^\text T)\mathbf K^{-1}
\end{equation}
where \(\mathbf X = [x,y,1]^\text T\) and \(\mathbf X' = [x',y',1]^\text T\) are the pixel locations in previous and current image respectively, $\mathbf H$ is the warp matrix, $\mathbf R$, $\mathbf t_0$ are the rotation matrix and translation vector from the second camera frame to the previous frame, $\mathbf t$ is the unscaled translation, $\mathbf n$, $d$ are the unit normal vector and distance to the ground plane in second camera frame, and $\mathbf K$ is the camera intrinsic matrix (assumed known).

During optimisation we parameterise \(\mathbf R\) as a Rodrigues vector \(\mathbf r = [r_x,r_y,r_z]^\text T\) and \(\mathbf n\) as~\cite{crispell2008g}
\begin{align}
	\theta & = \tan^{-1}(n_y/n_x) \\
	\phi   & = \sin^{-1}(n_z)     
\end{align}

Since the IMU provides reliable orientation information, especially for pitch and roll, out of the three possible parameterisations : \(\mathbf{p} = [t_x, t_y, t_z]^\text T\), \(\mathbf{p} = [t_x, t_y, t_z, r_x, r_y, r_z]^\text T\) and \(\mathbf{p} = [t_x, t_y, t_z, r_x, r_y, r_z, \theta, \phi]^\text T\), we choose the second since it provides the most accurate homography optimisation and tracking performance. The underlying assumption for fixing \(\mathbf n\) is that the ground is horizontal and therefore the normal vector depends only on the MAV's orientation. The validity of this assumption will be evaluated in Sec.~\ref{sec:evaluation}.

\subsection{Homography Estimation Cost Function}
The parameters of the warp matrix $\mathbf{H}$ are estimated by minimising the Sum of Squared Differences (SSD) error between image pixel intensities of the reference and warped images. However, a purely photometric cost minimisation may provide incorrect camera pose estimates due to a lack of observability or in the event of non-planar objects in the camera field of view. Since the IMU provides reliable orientation information, we add a penalty term which biases the homography solution and avoids these local minima.

Suppose $\mathbf X=T(\mathbf X';\mathbf p)$ stands for the homography mapping parameterised by the vector $\mathbf p$, we have
\begin{align}
	\mathbf p & =\underset{\mathbf p}{\argmin} \left(  f_{photo} + f_{imu} \right)  \nonumber           \\
	f_{photo} & = \sum_{j=1}^N \Vert I(T(\mathbf X_j';\mathbf p)) - I'(\mathbf X_j') \Vert ^2 \nonumber \\
	f_{imu}   & =  (\mathbf p-\mathbf p_0)^\text T\mathbf W(\mathbf p-\mathbf p_0)                      
\end{align}
where $\mathbf p_0$ is the initial guess obtained from IMU, $\mathbf W$ is a diagonal penalty weight matrix, $I$ and $I'$ are the previous and current image respectively, and $\mathbf X_j'$ is a pixel position in the evaluation region of current image.

\subsection{Gauss-Newton Optimisation}
We solve for the optimal parameters using iterative Gauss-Newton optimisation. After concatenating all the intensity values of pixel $X_j'$ in a vector, the Taylor expansion is
\begin{align}
	f(\mathbf p+\Delta \mathbf p)  = & \Vert \mathbf i(\mathbf p) + \mathbf G\Delta \mathbf p - \mathbf i' \Vert ^2 + \nonumber                              \\
	                                 & (\mathbf p+\Delta \mathbf p-\mathbf p_0)^\text T\mathbf W(\mathbf p+\Delta \mathbf p-\mathbf p_0) \label{eq:least_sq} 
\end{align}
where \( \mathbf i(\mathbf p) =  [I(T(\mathbf X_1';\mathbf p)), \dots, I(T(\mathbf X_N';\mathbf p))]^\text T\) and \(\mathbf i'  =  [I'(\mathbf X_1'), \dots, I'(\mathbf X_N')]^\text T\).
The iterative update to the parameter vector ends up being
\begin{equation}
	\Delta \mathbf p = \left(\mathbf G^\text T \mathbf G+\mathbf W \right)^{-1} \left(\mathbf G^\text T \left(\mathbf i'-\mathbf i(\mathbf p) \right) + \mathbf W(\mathbf p_0-\mathbf p) \right)
\end{equation}
where \(G\) is the Jacobian of the photometric residual term.
Note that as an implementation optimisation we only choose pixels with a high gradient magnitude similar to~\cite{shankar2016robust}. This significantly speeds up computation of the update with negligible loss in accuracy. The detailed timing performance is discussed in Sec.~\ref{sec:evaluation}.

%% file: ieeeconf/fusion.tex
\section{Visual-Inertial Fusion}
The optimisation in the previous section outputs an unscaled translation. Inspired by~\cite{grabe2013comparison}, we use an EKF to scale it to metric and additionally filter the frame-to-frame noise.

\subsection{Definition}
In the following section the superscripts and subscripts $C$ and $I$ imply a quantity in the camera and IMU frames respectively. The state vector contains camera velocity in the camera frame $^C \mathbf v$, distance to the plane from the camera $d$, and the linear acceleration bias in the IMU frame $^I \mathbf b$.
\[
	\mathbf x = [{^C \mathbf v ^\text T}, d, {^I \mathbf b ^\text T}]^\text T, \quad {^C \mathbf v}, {^I \mathbf b} \in \mathbb{R}^3, d \in \mathbb{R}
\]

\subsection{Prediction}
The derivative of $^C \mathbf v$ can be modeled~\cite{grabe2013comparison} as
\begin{align}
	^C \dot{\mathbf  v} & = {^C \mathbf R _I} \left( {^I \mathbf a} + [{^I \dot{\mathbf \omega}}]_\times {^I \mathbf p _{IC}}+[{^I \mathbf \omega _m}]_\times ^2 {^I \mathbf p _{IC}} \right) - [{^C \mathbf \omega _m}]_\times {^C \mathbf v} \nonumber \\
	                    & \approx {^C \mathbf R _I} \left( {^I \mathbf f _m} +{^I \mathbf g} +[{^I \mathbf \omega _m}]_\times ^2 {^I \mathbf p _{IC}} \right) - [{^C \mathbf \omega _m}]_\times {^C \mathbf v}                                           
\end{align}
where ${^C \mathbf R _I}$ is the rotation matrix from IMU frame to camera frame, ${^I \mathbf a}$ and ${^I \mathbf g}$ are the acceleration and gravity in the IMU frame, ${^I \mathbf f _m}$ and ${^I \mathbf \omega _m}$ are the raw linear acceleration and angular velocity measured by the IMU (subscript $m$ denotes raw measurement from visual odometry, IMU, or range finder), and ${^C \mathbf \omega}$ is the angular velocity in the camera frame. The subscript $\times$ denotes the skew symmetric matrix of the vector inside the bracket.

Therefore, the prediction process in discrete EKF can be written as
\begin{align}
	^{C}\hat{\mathbf v}{[k]_{{k-1}}} & = {}^{C}\hat{\mathbf v}[k-1] + \tau ~ {}^{C}\dot{\mathbf v}[k]                     \\
	\hat{d}[k]_{k-1}                 & = \hat{d}[k-1] + \tau ~ {}^{C}\hat{\mathbf{v}}[k-1]^{\text{T}}~ {}^C \mathbf{n}[k] \\
	^{I}\hat{\mathbf b}{[k]_{k-1}}   & = {}^{I}\hat{\mathbf b}{[k-1]}                                                     
\end{align}
where $\tau$ is the time step and ${^C \mathbf{n}}$ is the normal vector in camera frame (an alias for ${\mathbf{n}}$ defined in Sec. \ref{sec:homography_constraint}). The predicted states are denoted using $\hat{\mathbf x} {[k]_{k-1}}$.
Using the Jacobian matrix $\mathbf G[k]_{k-1}$ the predicted covariance matrix of system uncertainty $ \Sigma [k]_{k-1} \in \mathbb{R}^{7\times 7}$ is updated as
\begin{align}
	\Sigma{[k]_{k-1}} = \mathbf G \Sigma{[k]} \mathbf G^ \text T + \mathbf V
	\left[ \begin{matrix}
	\cov(^{I}\mathbf f_{m}) & \mathbf 0_{3\times 3}        \\
	\mathbf 0_{3\times 3}   & \cov(^{I}\mathbf \omega_{m}) 
	\end{matrix} \right]
	\mathbf V^T
\end{align}
where
\begin{align}
	\mathbf G & = \frac{\partial \hat{\mathbf x} [k]_{k-1}}{\partial \hat{\mathbf x} [k-1]}                                    \nonumber                                                                        \\
	& =\left[ \begin{matrix}
	\mathbf I_3 - \tau [^{C}\mathbf \omega]_\times & \mathbf 0_{3\times 1}                                                                           & -\tau\ ^{C}\mathbf R_{I}                                                                                                                     \\
	\tau \ ^{C}\mathbf n^\text T                   & 1                                                                                               & \mathbf 0_{3\times 1}                                                                                                                        \\
	\mathbf 0_{3\times 3}                          & \mathbf 0_{3\times 1}                                                                           & \mathbf I_3                                                                                                                                  
	\end{matrix} \right] \in \mathbb{R}^{7\times 7}\\
	\mathbf V                                      & = \left[ \begin{matrix}  \frac{\partial \hat{\mathbf x} [k]_{k-1}}{\partial  ^{I}\mathbf f_{m}} & \frac{\partial \hat{\mathbf x} [k]_{k-1}}{\partial  ^{I}\mathbf \omega_{m}} \end{matrix} \right]	                                  \nonumber \\
	& = \left[ \begin{matrix}
	\tau \ ^{C}\mathbf R_{I} & \tau \left( ^{C}\mathbf R_{I} \mathbf M + [^{C}\mathbf v]_{\times} {^{C}\mathbf R_{I}} \right) \\
	\mathbf 0_{1\times 3}    & \mathbf 0_{1\times 3}                                                                          \\
	\mathbf 0_{3\times 3}    & \mathbf 0_{3\times 3}
	\end{matrix} \right] \in \mathbb{R}^{7\times 6}\\
	\mathbf M & = \left( ^{I}\mathbf \omega_{m}^\text T \ {^{I}\mathbf p_{IC}} \right) \mathbf{I_3} + ^{I}\mathbf \omega_{m} \ {^{I}\mathbf p_{IC}^\text T} - 2{^{I}\mathbf p_{IC}} {^{I}\mathbf \omega_{m}^\text T}
\end{align}

\subsection{Update}
When both unscaled translation between two frames $\mathbf t_m[k]$  and range sensor signal $l_m[k]$ are available for update, the measurement vector $\mathbf z_m[k]$ is
\begin{align}
	\mathbf z_m[k]              & =	\left[ \begin{matrix}{\mathbf t_m[k] / \tau}                                    \\ l_m[k] n_{z_m}[k]\end{matrix} \right]
	\intertext{where the $n_{z_m}[k]$ is the z component of ${^C \mathbf{n}}$, and the subscript $m$ denotes direct measurements. The predicted measurement based on $\hat{\mathbf x} {[k]_{k-1}}$ is }
	\hat{\mathbf z} {[k]_{k-1}} & =	\left[ \begin{matrix}{{^{C}\hat{\mathbf v} ([k]_{k-1})} / {\hat d ([k]_{k-1})}} \\ {\hat d ([k]_{k-1})}	\end{matrix} \right]
\end{align}
Calculating the Kalman gain $\mathbf K [k] \in \mathbb{R}^{7\times 4}$
\begin{align}
	\mathbf K[k]        & = \Sigma J^\text T \left( \mathbf J \Sigma \mathbf J^\text T + \cov(\mathbf z_m) \right)^{-1}              \\
	\mathbf J           & = \left[ \begin{matrix}
	{\hat{\mathbf z} {[k]_{k-1}} }\over{^{C}\hat{\mathbf v} {[k]_{k-1}} } & {\hat{\mathbf z} {[k]_{k-1}} }\over{\hat d {[k]_{k-1}} }           & {\hat{\mathbf z} {[k]_{k-1}} }\over{^{I}\hat{\mathbf b} {[k]_{k-1}} } 
	\end{matrix} \right] \qquad \qquad \nonumber                                         \\
	& = \left[ \begin{matrix}
	{{1}\over{\hat d {[k]_{k-1}} } } \mathbf I_3                          & - {^{C}\hat{\mathbf v} {[k]_{k-1}} } \over {\hat d {[k]_{k-1}^2} } & \mathbf 0_{3\times 3}                                                 \\
	\mathbf 0_{1\times 3}                                                 & 1                                                                  & \mathbf 0_{1\times 3}                                                 
	\end{matrix} \right] \in \mathbb{R}^{4\times 7}
	\intertext{Estimates $\hat{\mathbf x} [k]$ and $\Sigma [k]$ are updated accordingly as}
	\hat{\mathbf x} [k] & = \hat{\mathbf x} {[k]_{k-1}} + \mathbf K[k] \left( \mathbf z_{m}[k] - \hat{\mathbf z} {[k]_{k-1}} \right) \\
	\Sigma[k]           & = \left( \mathbf I_7 - \mathbf K\mathbf J \right) \Sigma {[k]_{k-1}}
\end{align}

%% file: ieeeconf/simulation_experiment.tex
\section{Evaluation}\label{sec:evaluation}
We evaluate the performance of our approach on a wide variety of scenarios and compare and contrast performance with state of the art algorithms. We first present experimental setup and results in simulation followed by those with real-world data obtained from an aerial platform.

\subsection{Benchmarks and Metrics}
Our method (RaD-VIO) is compared to the tracker proposed in \cite{grabe2015nonlinear} (Baseline), for which we implement the optical flow method described in~\cite{grabe2012board} and, for fair comparison, use the same EKF fusion methodology as our approach. The resulting tracking errors of the EKF with a range finder are much smaller than those when using the EKF in \cite{grabe2013comparison}. We choose this as the baseline since it is also based on the homography constraint and assumes local planarity. Additionally, we also compare with a state-of-the-art monocular visual-inertial tracker VINS-Mono~\cite{qin2018vins} without loop closure (VINS-D, VINS-downward). 

\begin{figure}[h]
	\begin{center}
		\includegraphics[scale = 0.5]{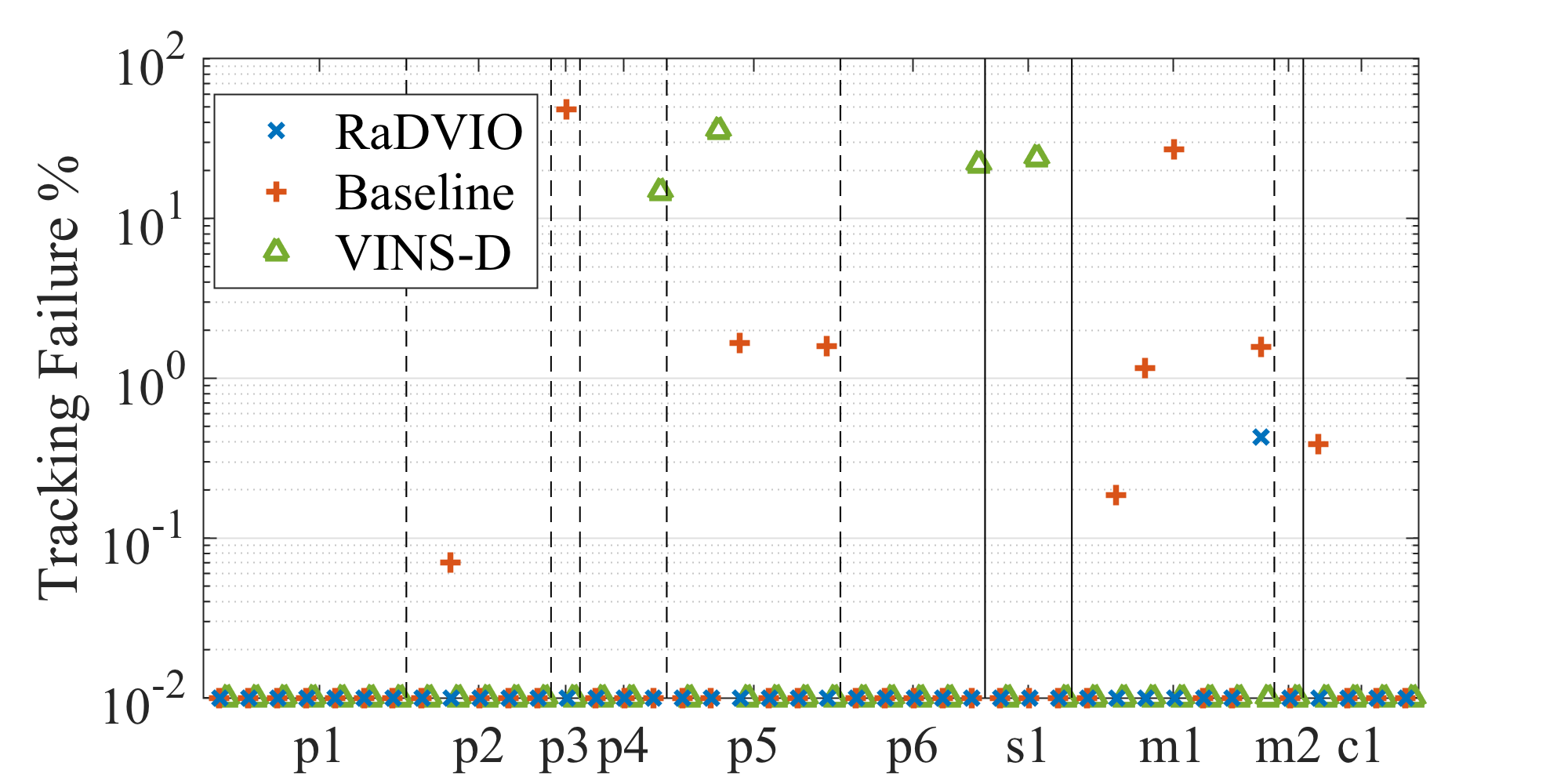} 
		\caption{Percentage of tracking failures: The failure instances of Baseline and RaD-VIO are calculated by a per-frame counter, while that of VINS-D is estimated according to gap between messages, not including time taken for initialisation. } 
		\label{fig:failures}
	\end{center}
\end{figure}

The metrics used are Relative Pose Error (RPE) (the interval is set to 1s) and Absolute Trajectory Error (ATE)~\cite{kummerle2009measuring}. For ATE, we only report results in the xy plane since the altitude is directly observed by the rangefinder. We also report the number of times frame-to-frame tracking fails in Fig.~\ref{fig:failures}. Since  RaD-VIO and Baseline output velocity, the position is calculated using dead-reckoning. Since a lot of our trajectories are not closed loops, instead of reporting ATE we divide it by the cumulative length of the trajectory (computed at 1s intervals) in the horizontal plane to get the relative ATE. We try to incorporate an initial linear movement in test cases to initialise VINS-D well, but a good initialisation is not guaranteed. For error calculation of VINS-D, we only consider the output of the tracker after it finishes initialisation.

\subsection{Simulation Experiments}
We utilise AirSim~\cite{airsim2017fsr}, a photorealistic simulator for aerial vehicles for the purpose of evaluation. The camera generates $240\times 320$ images at a frame rate of 80 Hz. The focal length is set to 300 pixels. The IMU and the single beam laser rangefinder output data at 200 Hz and 80 Hz respectively, and no noises are added.

\subsubsection{Simulation Test Cases}\label{sec:sim_setup}
\begin{figure}[h]
	\centering
	\begin{subfigure}[b]{.35\linewidth}
		\centering
		\includegraphics[clip, trim = 50 0 50 30, width=\textwidth]{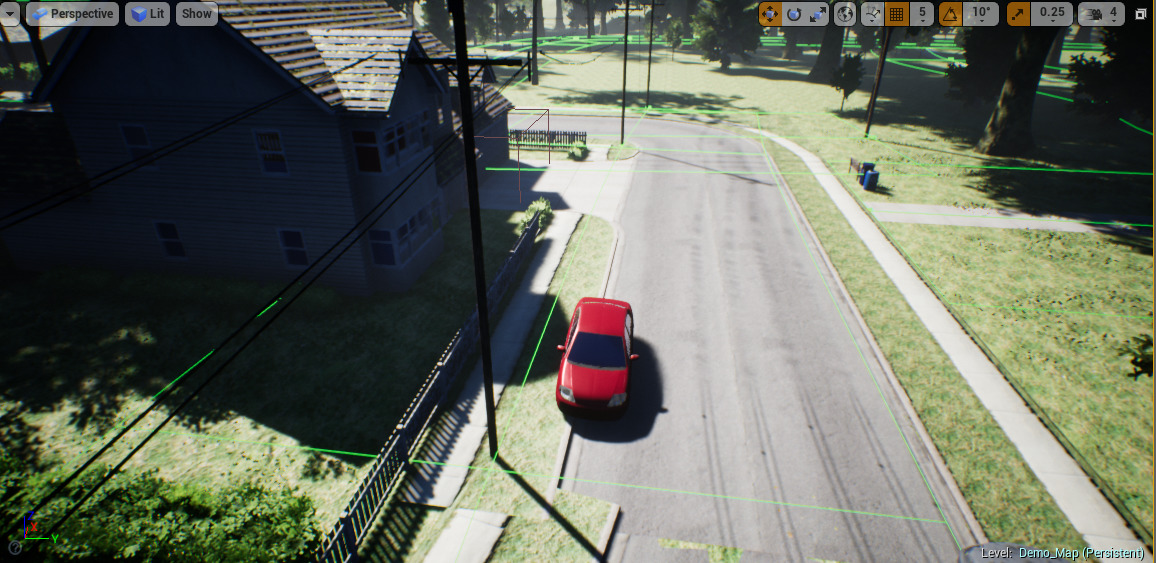}
	\end{subfigure}%
	\begin{subfigure}[b]{0.24\linewidth}
		\centering
		\includegraphics[width=0.48\textwidth]{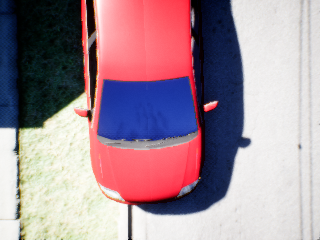}\hfil
		\includegraphics[width=0.48\textwidth]{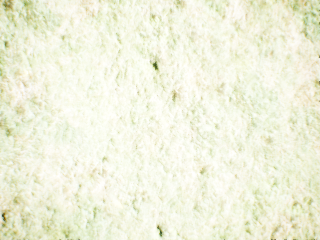}\\[0.5mm]
			\includegraphics[width=0.48\textwidth]{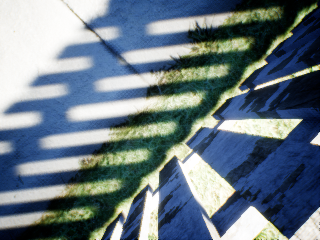}\hfil
			\includegraphics[width=0.48\textwidth]{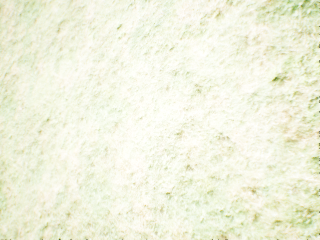}
		\end{subfigure}%
		\begin{subfigure}[b]{.35\linewidth}
			\centering
			\includegraphics[clip, trim = 50 0 50 30, width=\textwidth]{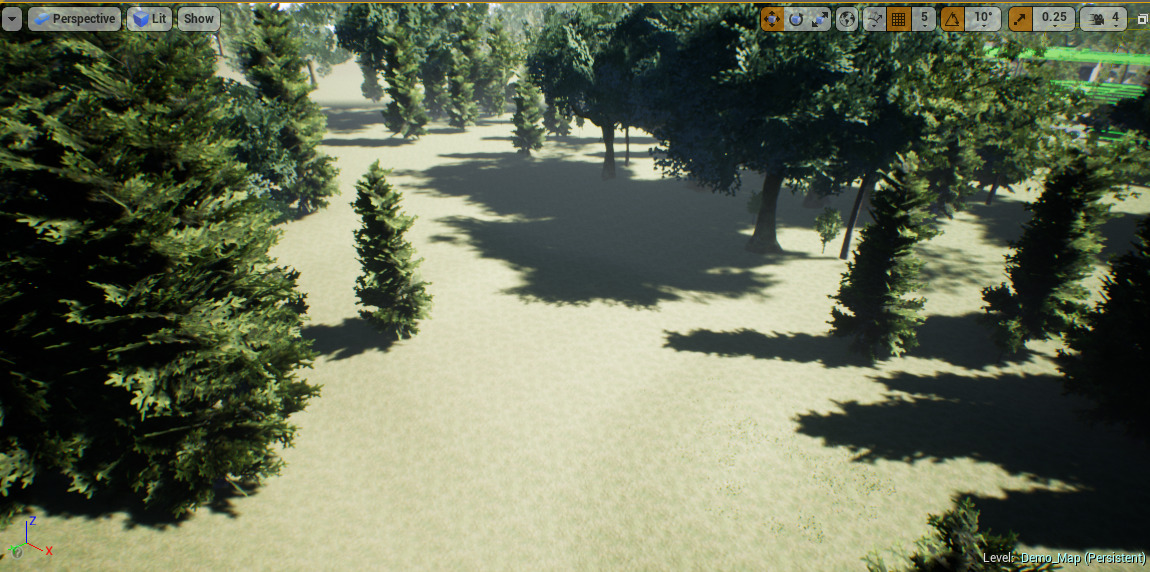}
		\end{subfigure}
												
		\begin{subfigure}[b]{.35\linewidth}
			\centering
			\includegraphics[clip, trim = 50 0 50 30, width=\textwidth]{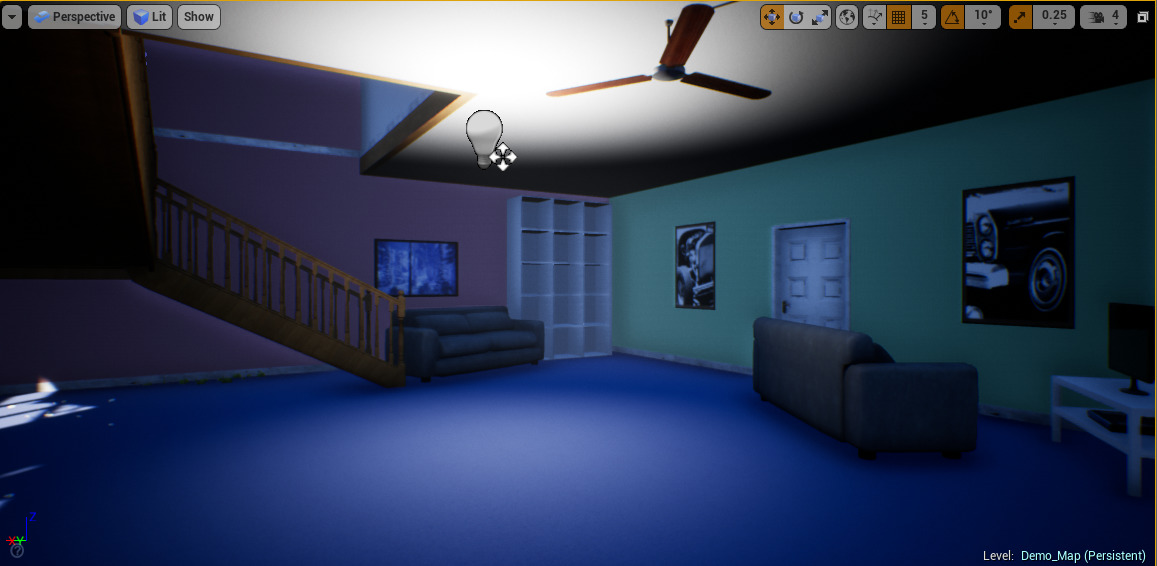}
		\end{subfigure}%
		\begin{subfigure}[b]{0.24\linewidth}
			\centering
			\includegraphics[width=0.48\textwidth]{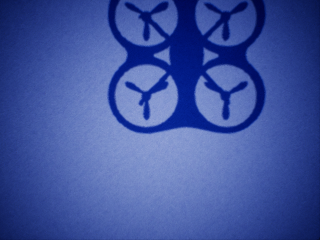}\hfil
			\includegraphics[width=0.48\textwidth]{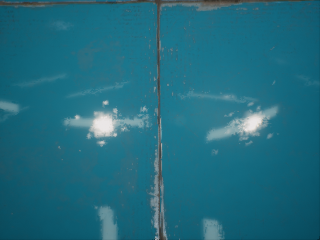}\\[0.5mm]
				\includegraphics[width=0.48\textwidth]{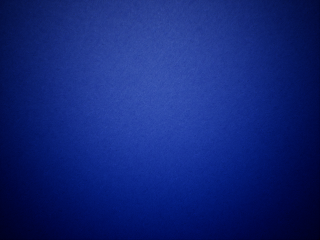}\hfil
				\includegraphics[width=0.48\textwidth]{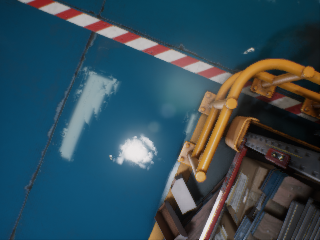}
			\end{subfigure}%
			\begin{subfigure}[b]{.35\linewidth}
				\centering
				\includegraphics[clip, trim = 50 0 50 10, width=\textwidth]{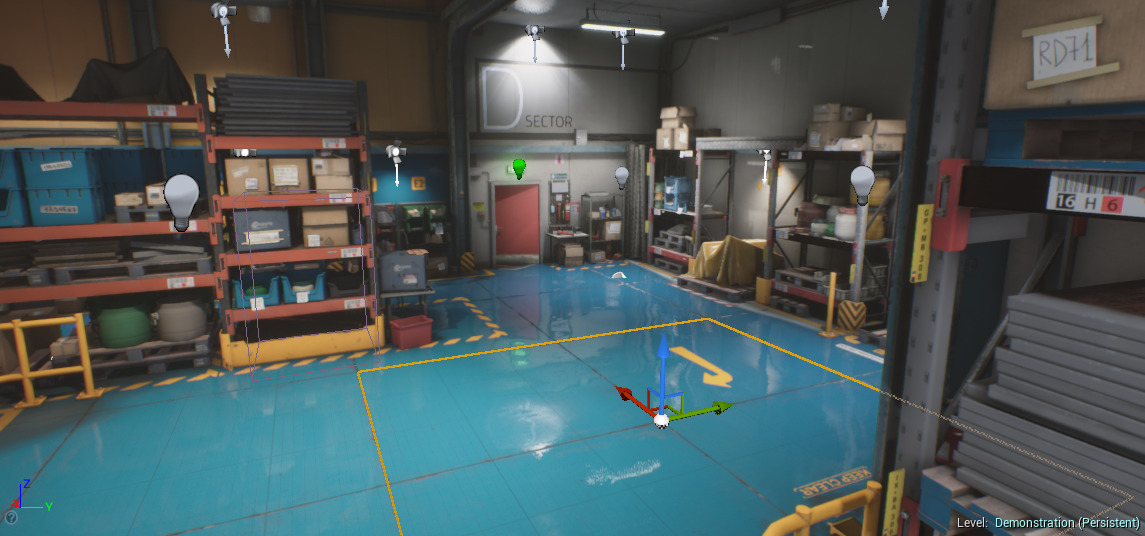}
			\end{subfigure}
			\caption{Sample views of environments used in the simulation experiments defined in Sec.~\ref{sec:sim_setup}}
			\label{fig:filmstrip}
		\end{figure}

For the tracking system to work, the following assumptions or conditions should be met or partly met:
\begin{itemize}
	\item The ground should be planar (homography constraint)
	\item It should be horizontal (parameterisation choice)
	\item The scene should be static (constancy of brightness)
	\item Small inter-frame displacement (applicability of warp).
\end{itemize}
Therefore, the following ten test scenarios are designed to evaluate the algorithms:
\begin{itemize}
	\item \textbf{p1}: All assumptions met - ideal conditions
	\item \textbf{p2}: Viewing planar ground with low texture
	\item \textbf{p3}: Viewing planar ground with almost no texture
	\item \textbf{p4}: Viewing planar ground with moving features
	\item \textbf{p5}: Vehicle undergoing extreme motion
	\item \textbf{p6}: Camera operating at a low frame rate
	\item \textbf{s1}: Viewing a sloped or a curved surface
	\item \textbf{m1}: Viewing a plane with small clutter 
	\item \textbf{m2}: Viewing moving features with small clutter
	\item \textbf{c1}: Viewing a plane with large amounts of clutter
\end{itemize}
Each of these cases are tested in diverse environments including indoors, road, and woods. We use 42 data sets in total for evaluation. Fig.~\ref{fig:filmstrip} shows some typical images from these datasets.

\subsubsection{Simulation Results and Discussion}
\begin{figure}[h]
	\captionsetup[subfigure]{labelformat=empty}
	\centering
	\begin{subfigure}[b]{\linewidth}
		\includegraphics[width=\linewidth]{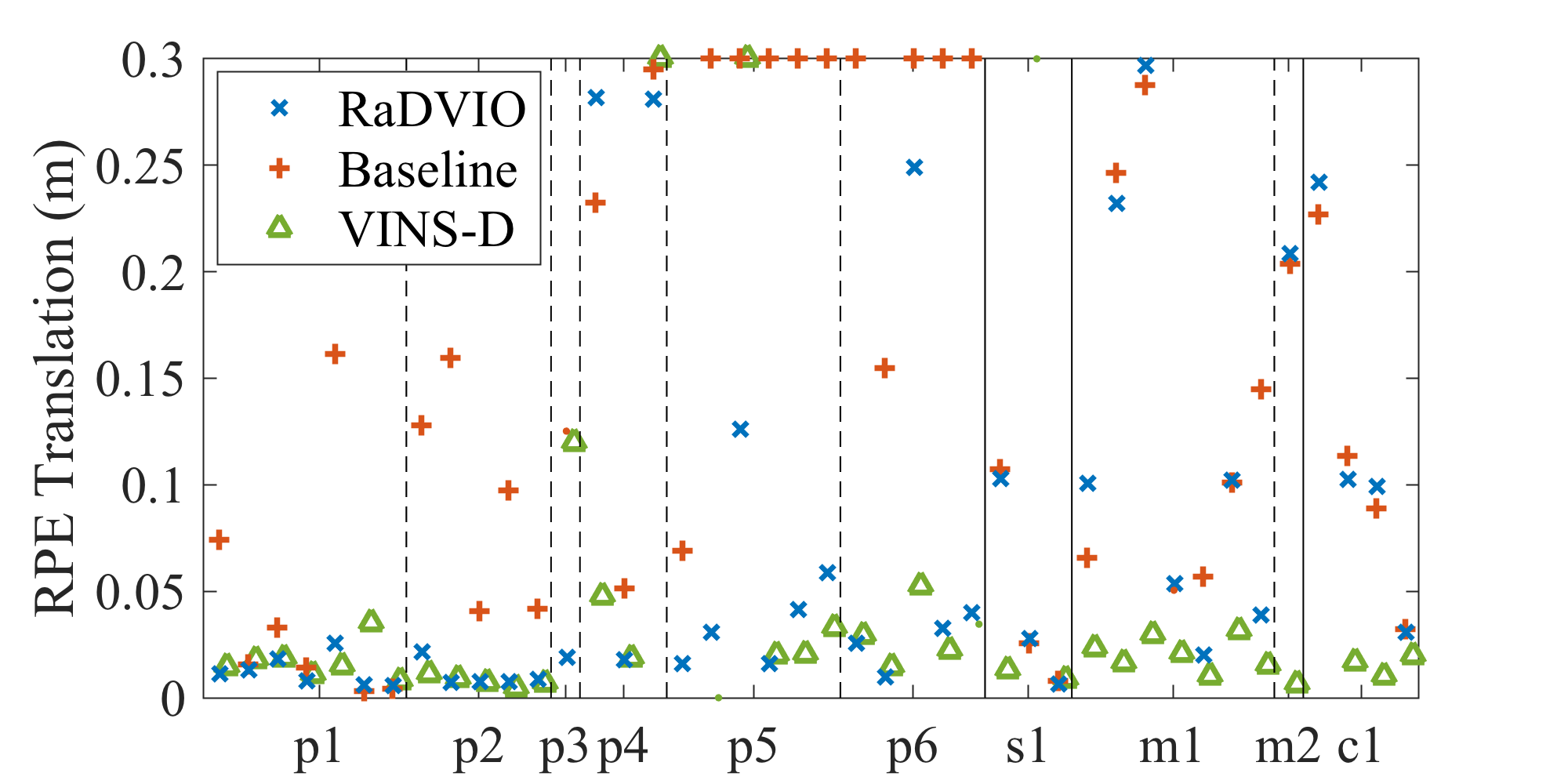} \caption{}
	\end{subfigure}
	\begin{subfigure}[b]{\linewidth}
		\includegraphics[width=\linewidth]{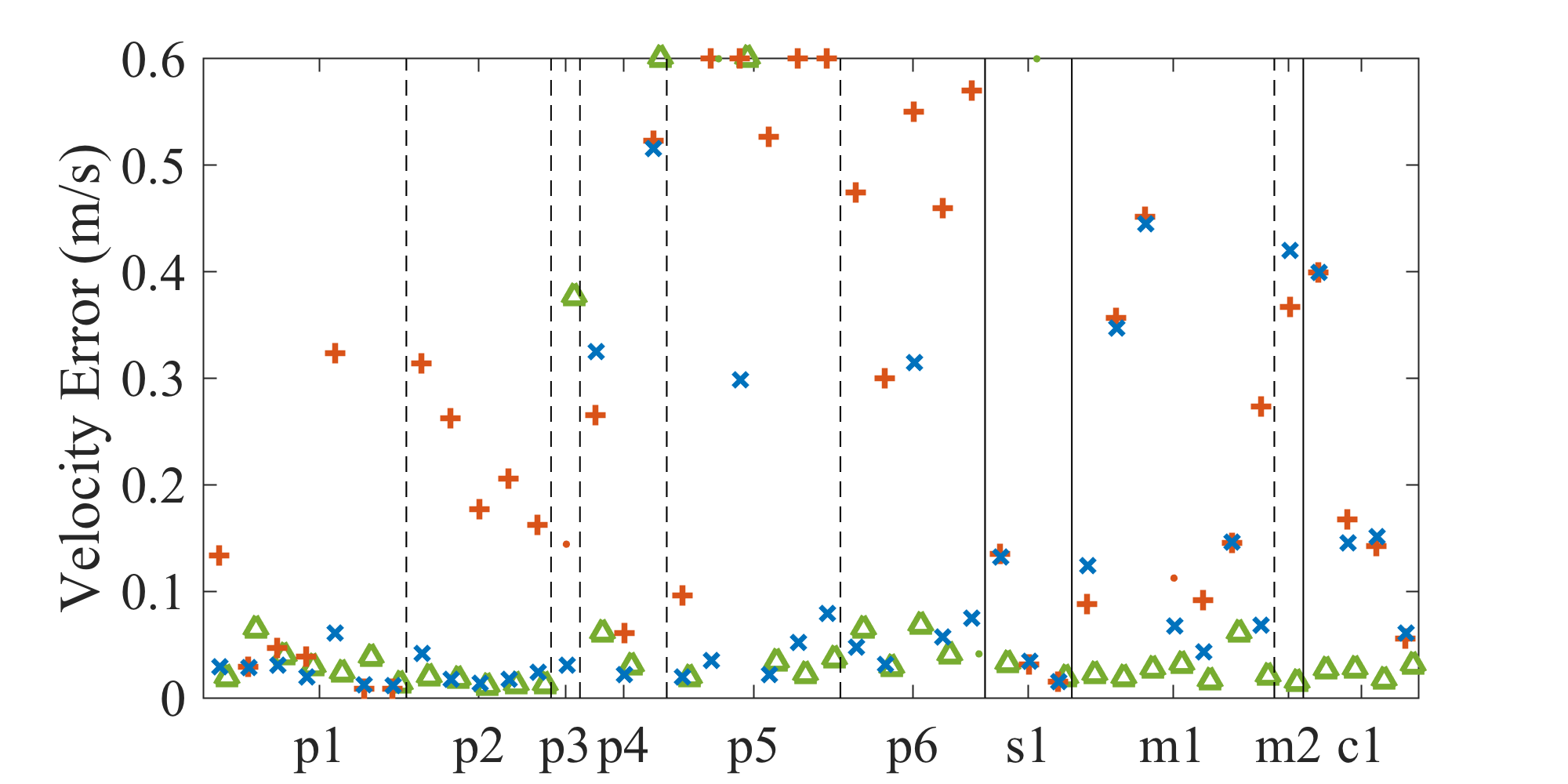} \caption{}
	\end{subfigure}
	\begin{subfigure}[b]{\linewidth}
		\includegraphics[width=\linewidth]{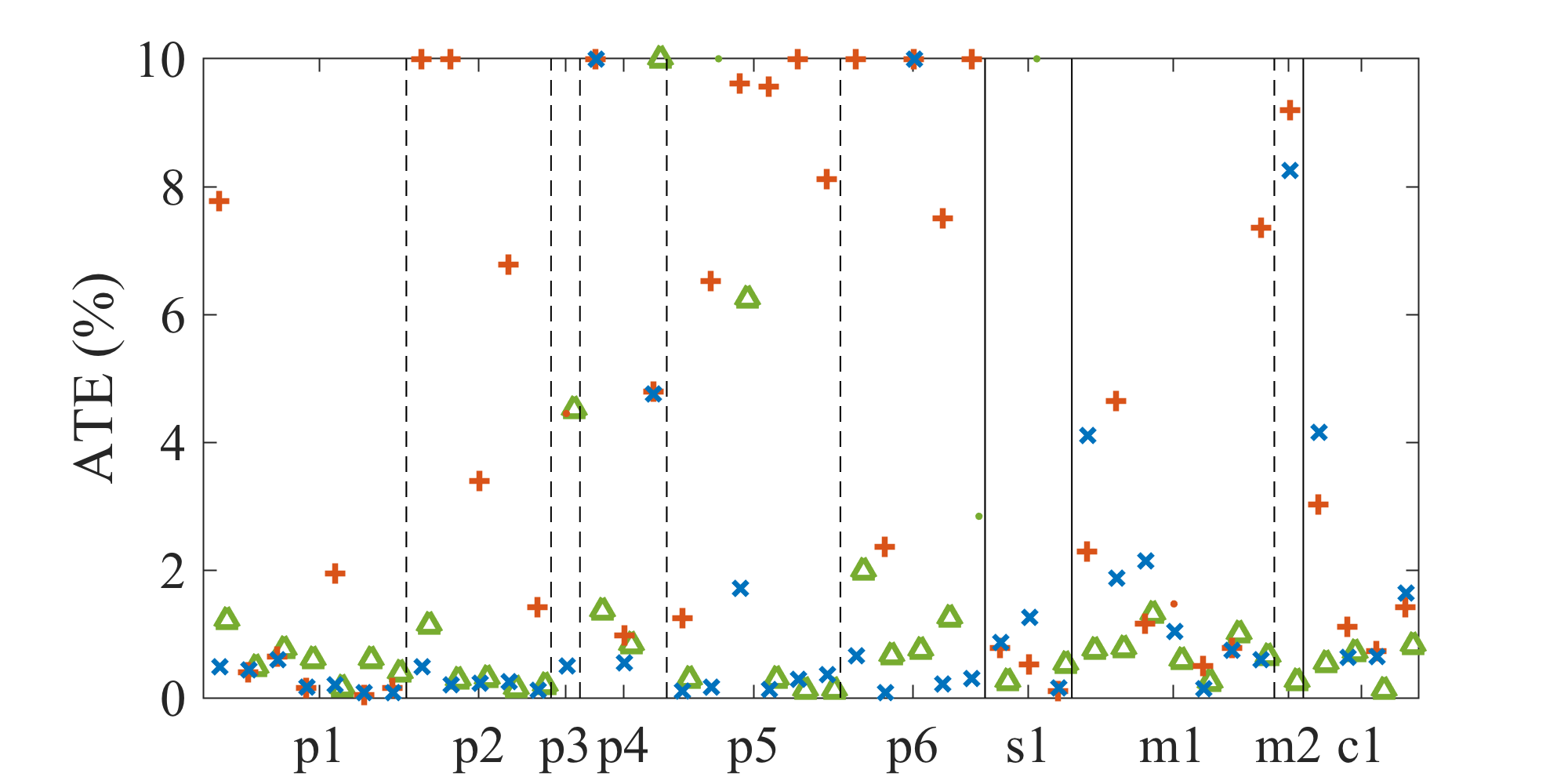} \caption{}
	\end{subfigure}
	\caption{Comparison of the algorithms on the simulation datasets described in Sec.\ref{sec:sim_setup}. Solid lines separate the cases where the planar assumption is met, not met or it is a slope, and dashed lines separates other assumptions. The point marker means the data point is not reliable due to a large number of tracking failures. The errors drawn at the upper boundary are clipped.}
	\label{fig:error_sim}
\end{figure}

The RPE and relative ATE of all test cases are shown in Fig.~\ref{fig:error_sim}. 
For relative ATE our method outperforms both Baseline and VINS-D in almost all the test cases. The velocity error in Fig.~\ref{fig:error_sim}~(c) contains both velocity bias and random error, and compared to VINS-D our method generates similar velocity errors when the planar assumption is satisfied, but larger when it is not. This error is much smaller than Baseline. 
												
\begin{figure}[h]
	\begin{subfigure}[b]{\linewidth}
	\centering
	\includegraphics[width=0.9\linewidth]{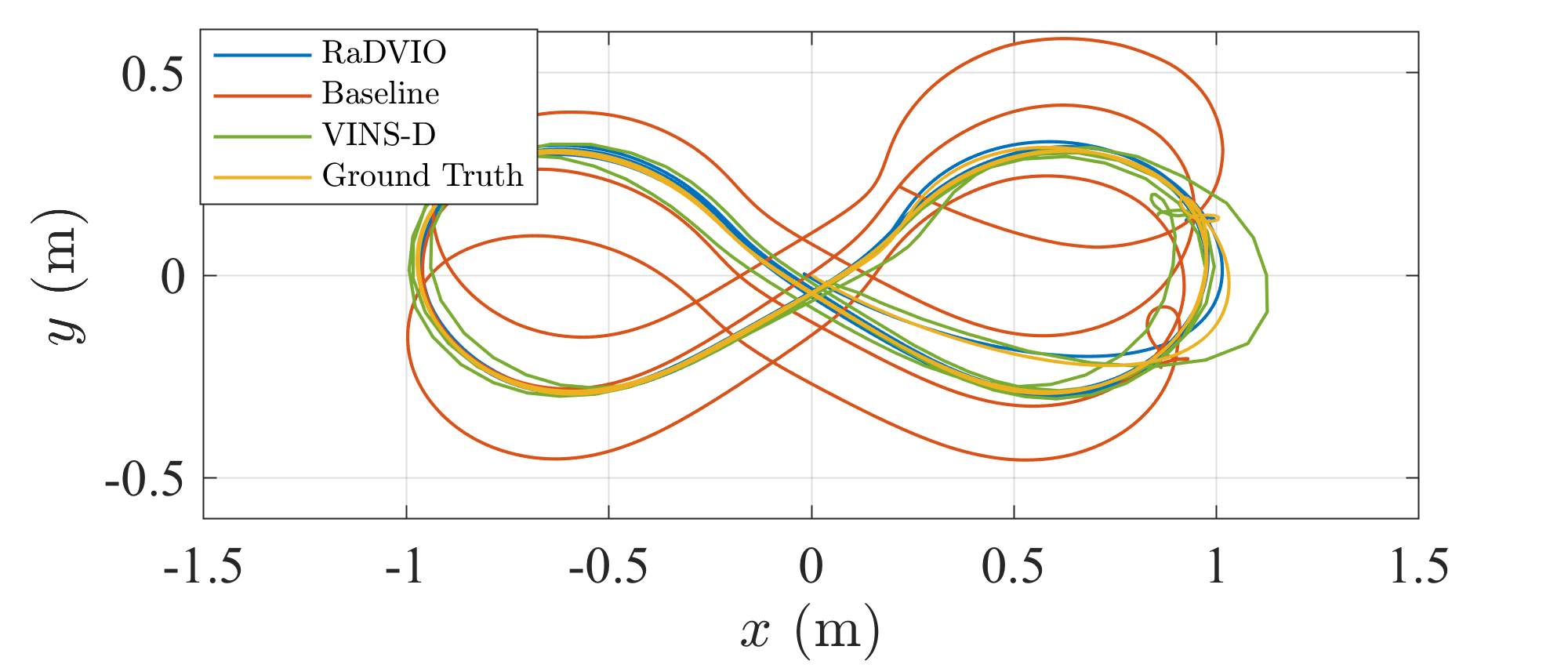}
		\caption{Dead-reckoned odometry tracks (aligned according to the calculation of ATE).}
	\end{subfigure}
	\begin{subfigure}[b]{\linewidth}
	\centering
		\includegraphics[width=0.9\linewidth]{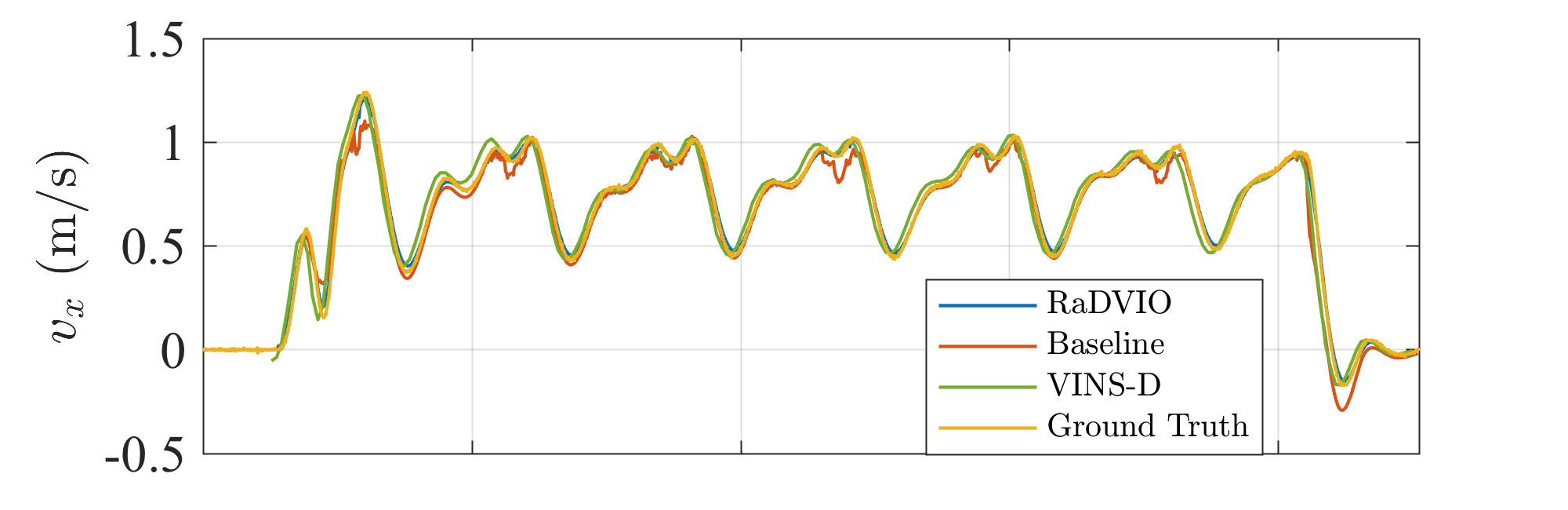}
		\includegraphics[width=0.9\linewidth]{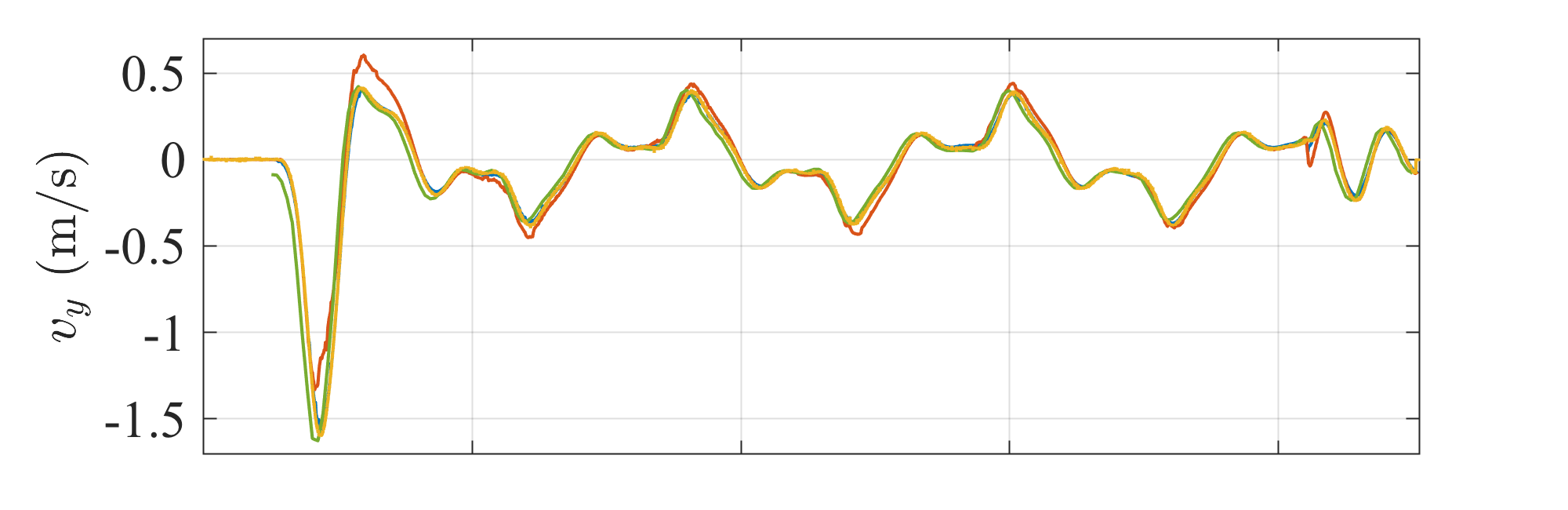}
		\includegraphics[width=0.9\linewidth]{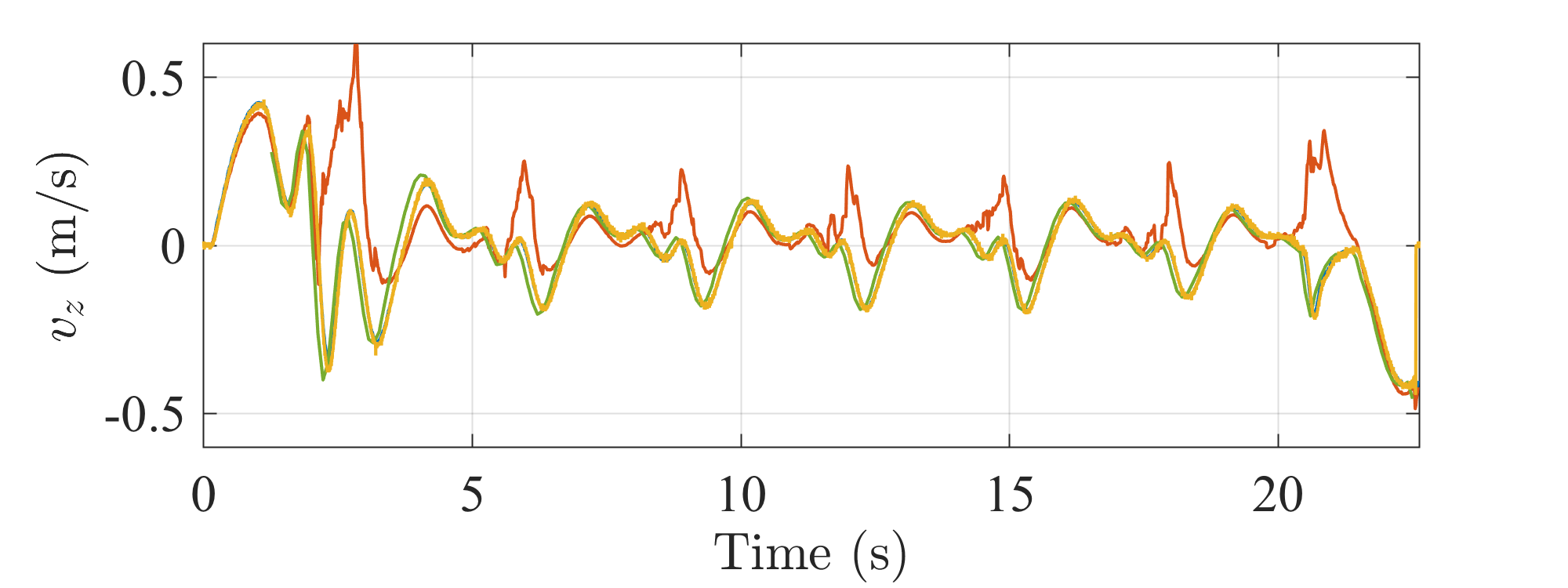} \\
		\caption{Corresponding velocities in MAV body frame.}
	\end{subfigure}
	\caption{Comparison of algorithms in simulation for a figure 8 trajectory.}
	\label{fig:fig_8}
\end{figure}

Sometimes, VINS-D takes a lot of time to initialise the system, and doesn't do well right after initialisation.
During the test, VINS-D occasionally outputs extremely wrong tracking results, and it has difficulty in initialising/reinitialising through three test cases (challenging lighting condition, too few texture, or extreme motion). In contrast our tracker never generates any extreme results due to implicitly being constrained by the frame-to-frame displacement. 
												
The simulation shows that RaD-VIO is overall more accurate than compared to Baseline, and when all the assumptions are met, it is slightly more accurate than VINS-D. The test cases demonstrate that the proposed tracker is robust to a lot of non-ideal conditions.
												
\subsection{Real-world Experiments}
												
\subsubsection{Setup and Test Cases} \label{sec:real_setup}
To verify the real-world performance of our proposed approach, indoor and outdoor data was collected. The indoor data was obtained from a custom hexrotor platform in a motion capture arena. We use a MatrixVision BlueFox camera configured to output $376 \times 240$ resolution images at a frame rate of 60 Hz, and a wide angle lens with a focal length of 158 pixels. The frame rates of the TeraRanger One rangefinder and the VN-100 IMU are 300 Hz and 200 Hz respectively. The standard deviation of IMU angular velocity and linear acceleration are $0.02 \ \text{rad/s}$ and $1 \ \text{m/s}^2$ respectively. The indoor ground truth is provided by a VICON motion capture system. For outdoor data, only GPS data is provided as a reference, and we use an Intel Aero Drone\footnote{\url{https://www.intel.com/content/www/us/en/products/drones/aero-ready-to-fly.html}} for data collection. The same algorithm parameters as in the simulation experiments were used for the vision and optimisation frontends, and the parameters for the EKF were tuned based on the sensor noise for the respective configuration.
												
We evaluate 10 flight data sets collected according to the same classification criteria. The severity of moving features, and the platform motion is comparatively more significant in the corresponding real data sets. Note that the abbreviation of the test cases have the same meaning as in simulation section we mentioned before.
												
\begin{figure}[ht]
	\centering
	\captionsetup[subfigure]{labelformat=empty}
																				
	\begin{subfigure}[b]{.3\linewidth}
		\centering
		\includegraphics[width=\textwidth]{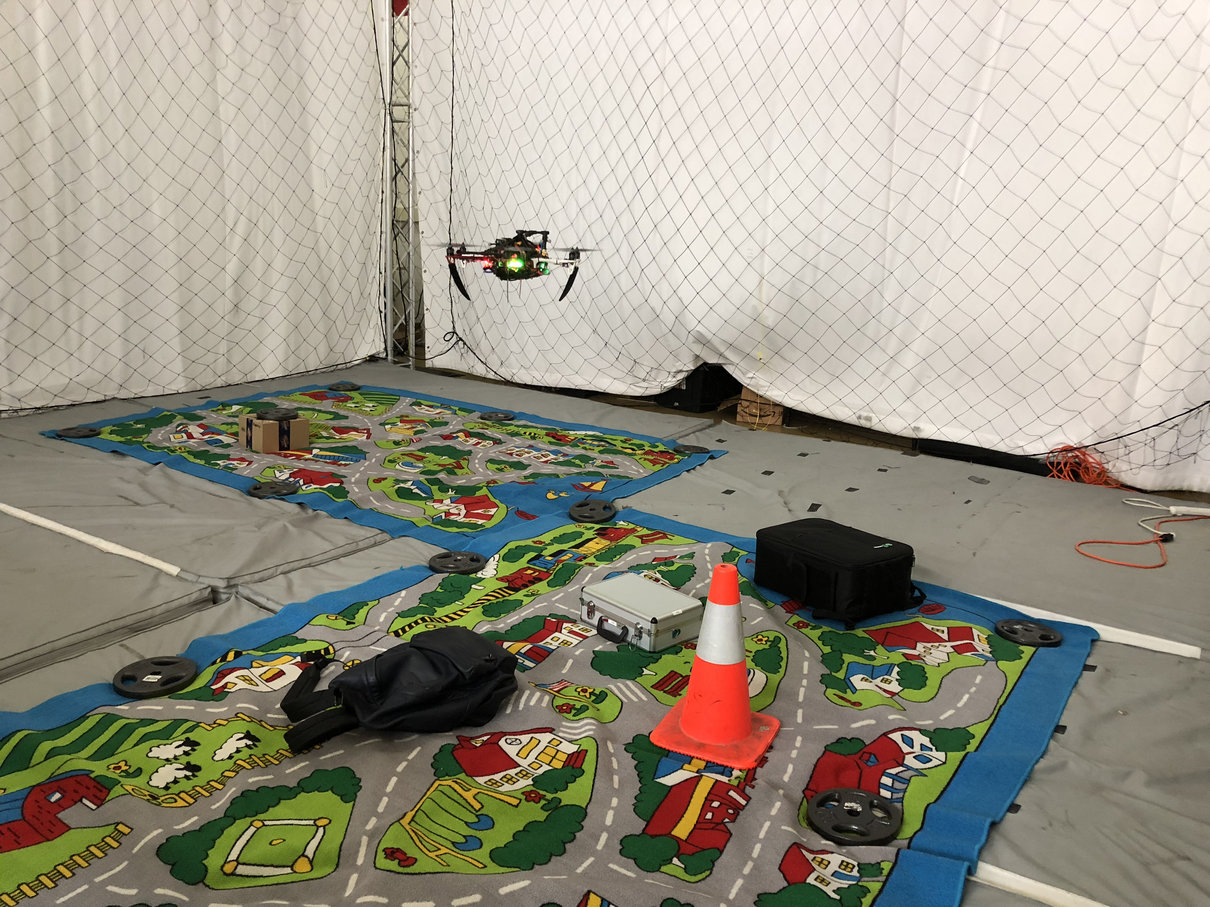}
	\end{subfigure}%
	\begin{subfigure}[b]{.3\linewidth}
		\centering
		\includegraphics[clip, trim = 27 4 27 4,width=0.48\textwidth]{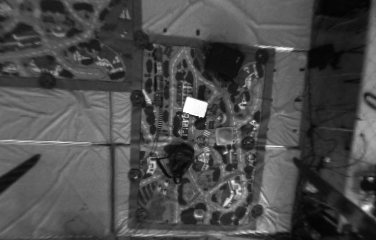}\hfil
		\includegraphics[clip, trim = 27 4 27 4,width=0.48\textwidth]{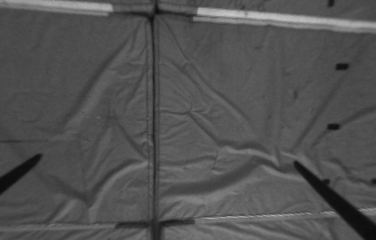}\\[0.5mm]
		\includegraphics[clip, trim = 27 4 27 4,width=0.48\textwidth]{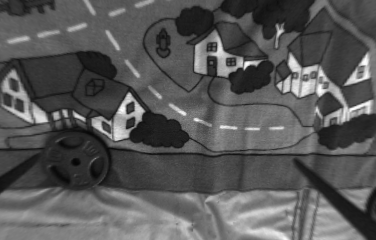}\hfil
		\includegraphics[clip, trim = 27 4 27 4,width=0.48\textwidth]{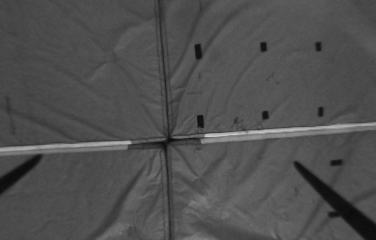}
		\end{subfigure}%
		\begin{subfigure}[b]{.3\linewidth}
			\centering
			\includegraphics[width=\textwidth]{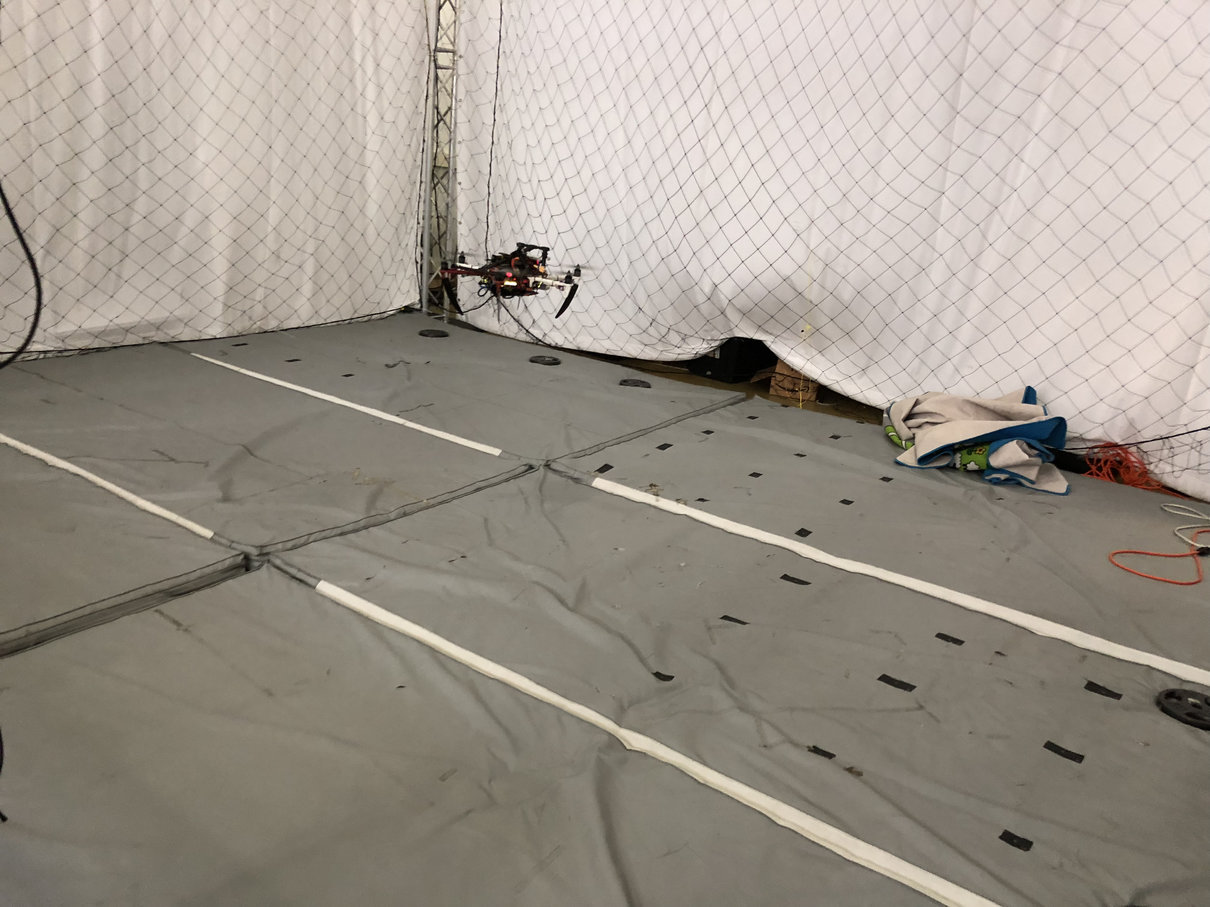}
		\end{subfigure}
																										
		\begin{subfigure}[b]{.3\linewidth}
			\centering
			\includegraphics[width=\textwidth]{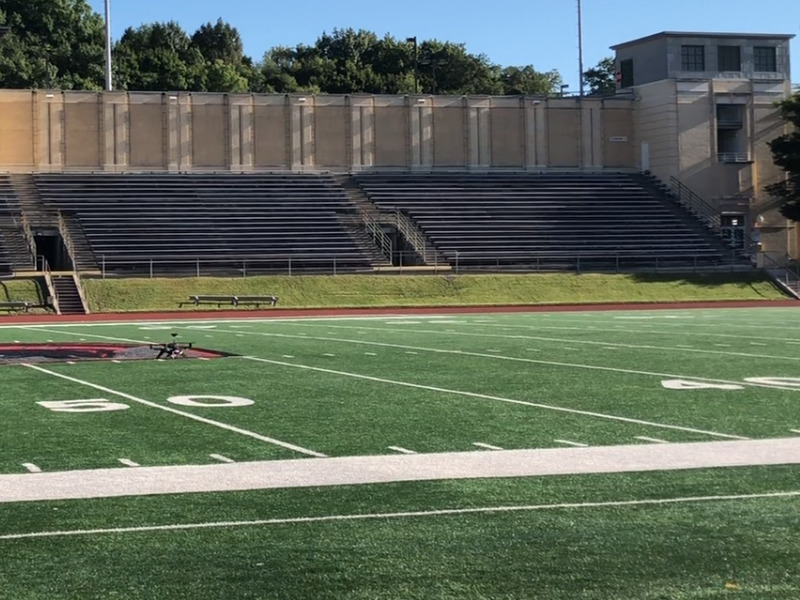}
		\end{subfigure}%
		\begin{subfigure}[b]{.3\linewidth}
			\centering
			\includegraphics[width=0.48\textwidth]{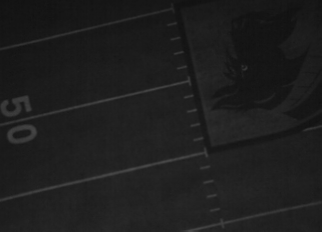}\hfil
			\includegraphics[width=0.48\textwidth]{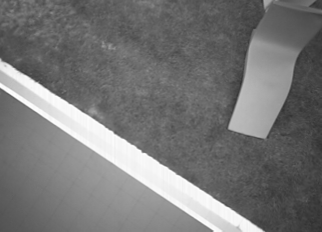}\\[0.5mm]
				\includegraphics[width=0.48\textwidth]{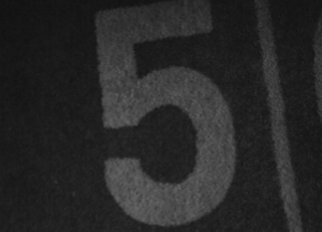}\hfil
				\includegraphics[width=0.48\textwidth]{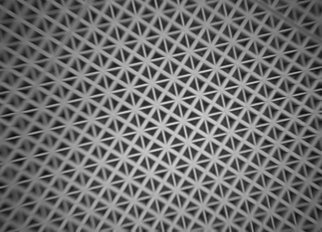}
			\end{subfigure}%
			\begin{subfigure}[b]{.3\linewidth}
				\centering
				\includegraphics[width=\textwidth]{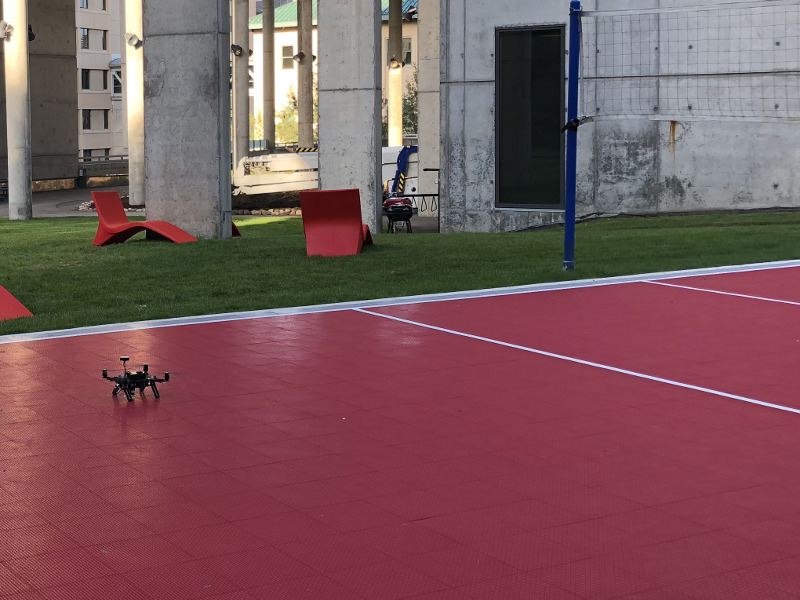}
			\end{subfigure}
		\caption{Sample views of environments used in the real-world experiments discussed in Sec.~\ref{sec:real_setup}}
			\label{fig:real_filmstrip}
		\end{figure}

\subsubsection{Experiment Results and Discussion}
\begin{figure}[h]
	\captionsetup[subfigure]{labelformat=empty}
	\centering
	\begin{subfigure}[b]{\linewidth}
		\includegraphics[width=\linewidth]{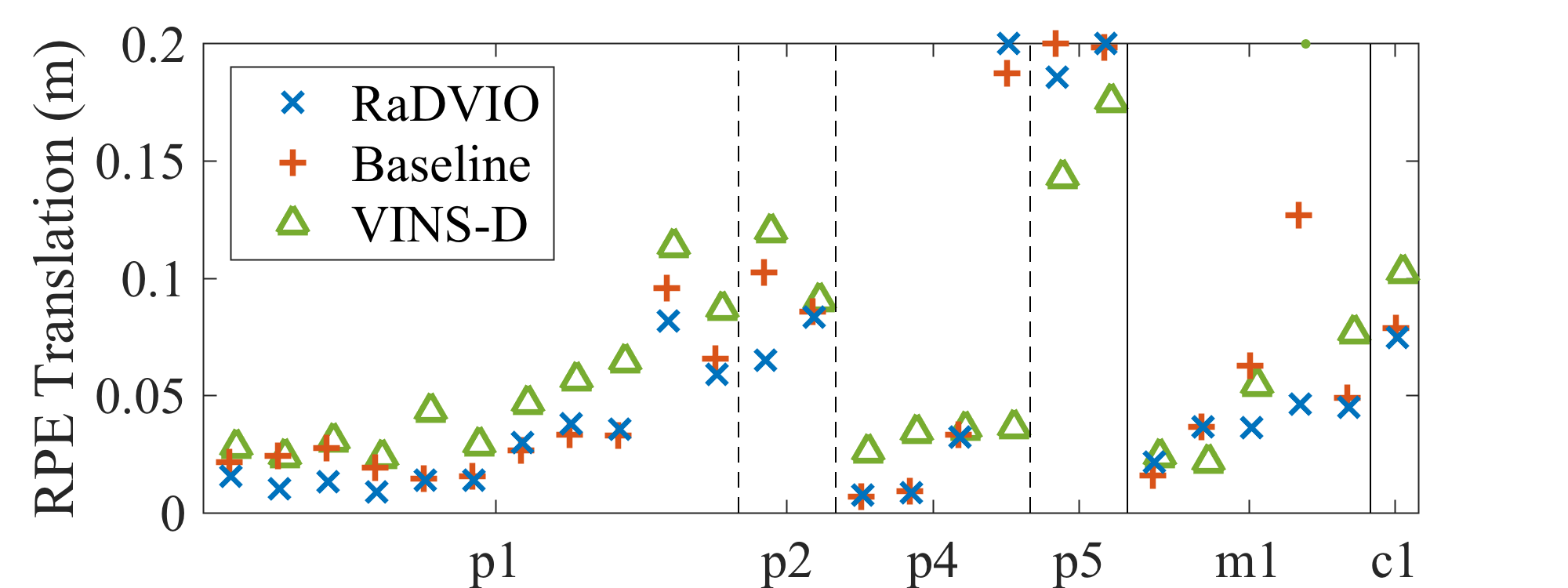} \caption{}
	\end{subfigure}
	\begin{subfigure}[b]{\linewidth}
		\includegraphics[width=\linewidth]{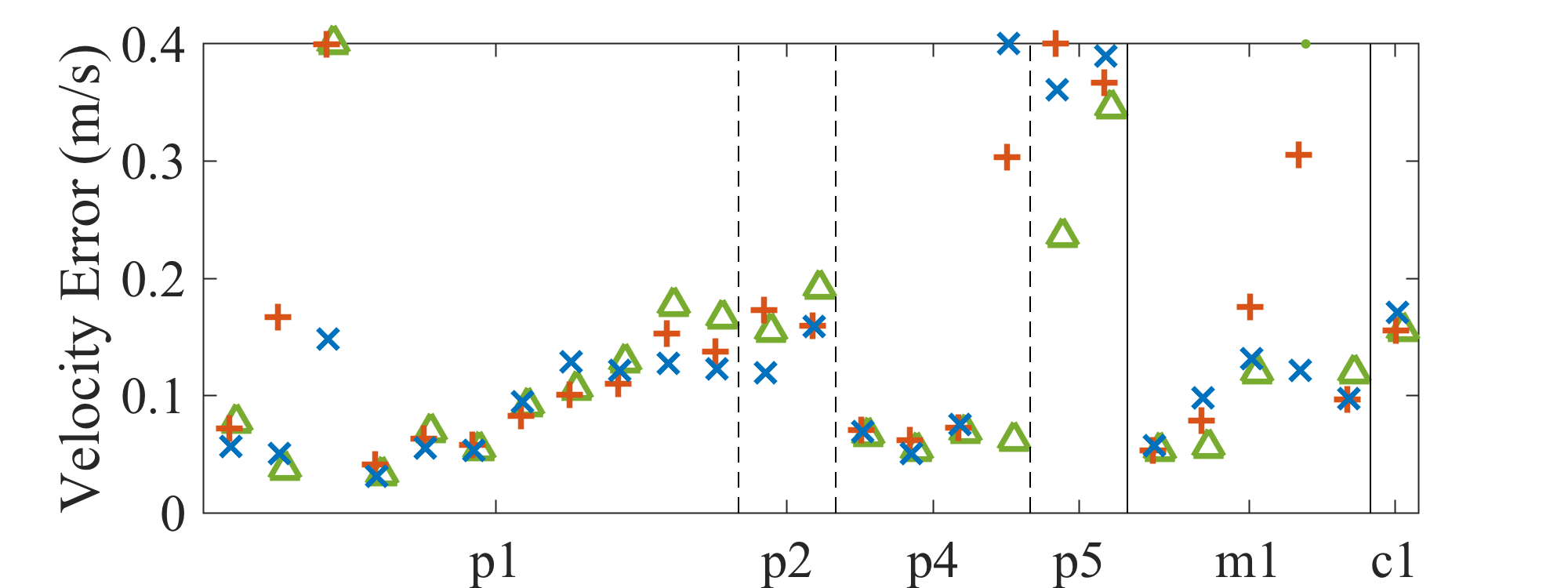} \caption{}
	\end{subfigure}
	\begin{subfigure}[b]{\linewidth}
		\includegraphics[width=\linewidth]{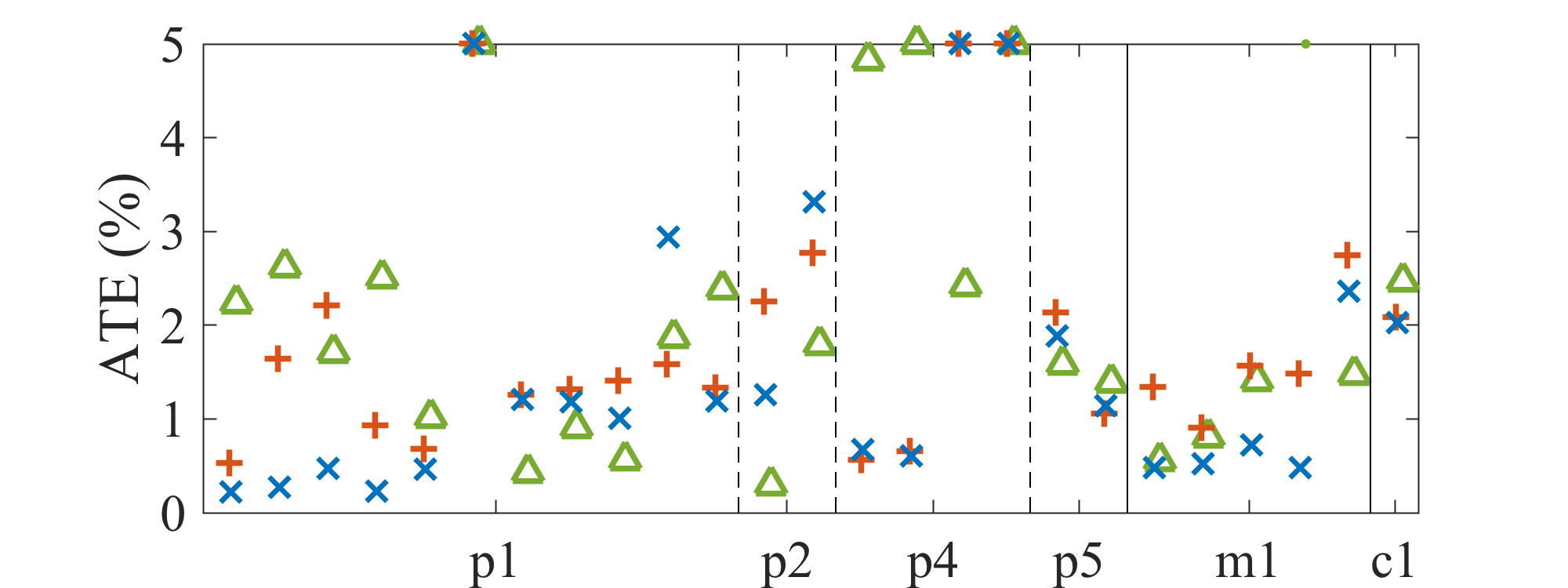} \caption{}
	\end{subfigure}
	\caption{Comparison of the algorithms on the real-world datasets described in Sec.\ref{sec:real_setup}. Solid lines separate the cases where the planar assumption is met, not met or it is a slope, and dashed lines separates other assumptions. The point marker means the data point is not reliable due to a large number of tracking failures. The errors drawn at the upper boundary are clipped.}
	\label{fig:error_real}
\end{figure}

The tracking errors are shown in Fig.~\ref{fig:error_real}. Similar to the simulation results for relative ATE our tracker is on average better than both Baseline and VINS-D. As for the translation part of RPE, our method is better or no worse than the other two methods except for the last case in p4 (moving features) and p5 (extreme motion). In these cases the robust sparse feature selection in VINS-D avoids being overly influenced. RaD-VIO is not as robust to extreme motion as it is in simulation, and there are two reasons: the IMU input is more noisy and the wider field of view ends up capturing objects outside the ground plane that adversely affect the alignment. For the rotation component of RPE, both Baseline and RaD-VIO are better than VINS-D; this is because both approaches use the IMU rotation directly.
																										
\begin{figure}[h]
	\centering
		\includegraphics[width=0.9\linewidth]{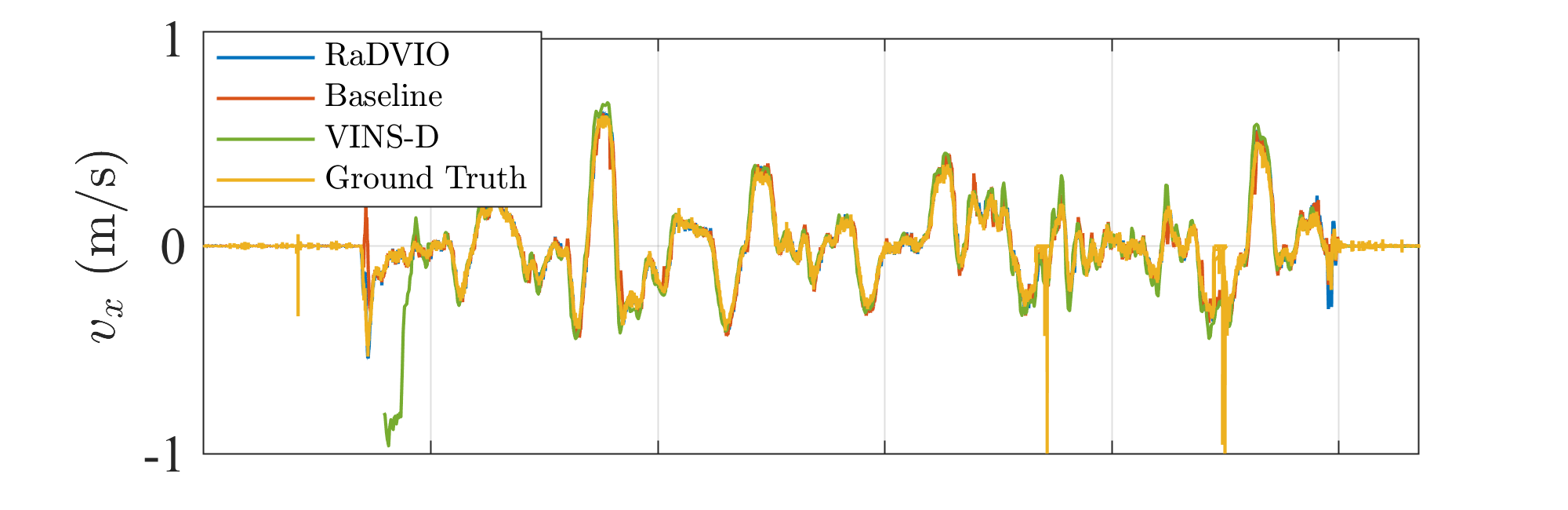}
		\includegraphics[width=0.9\linewidth]{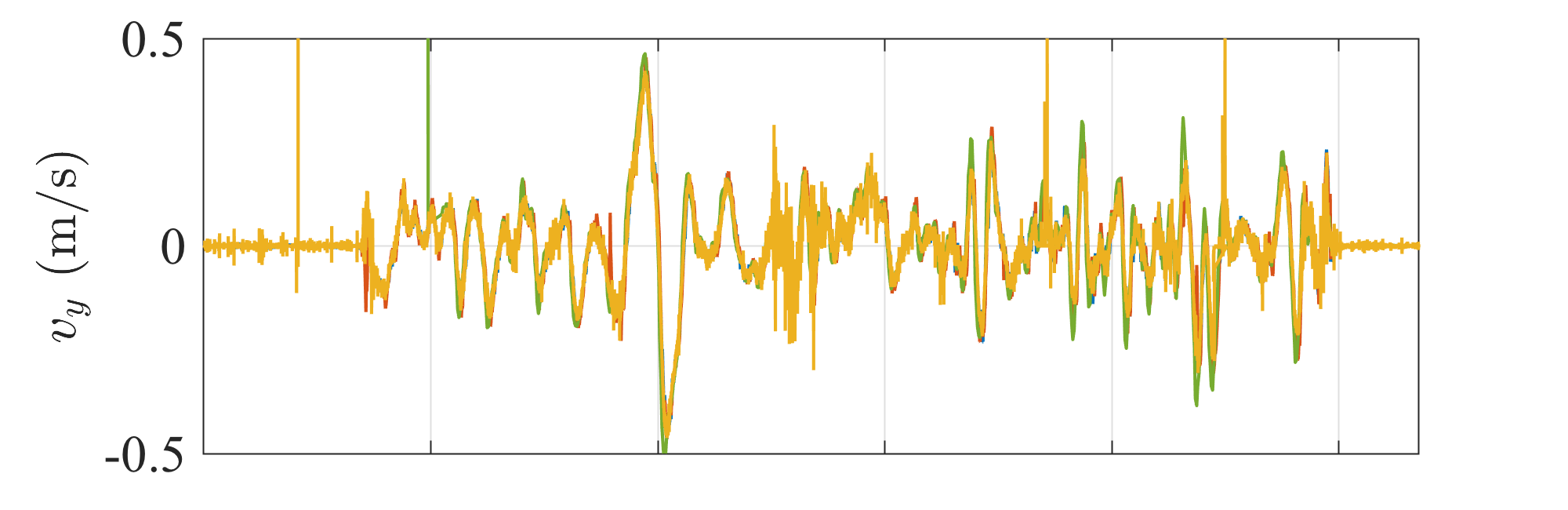}
		\includegraphics[width=0.9\linewidth]{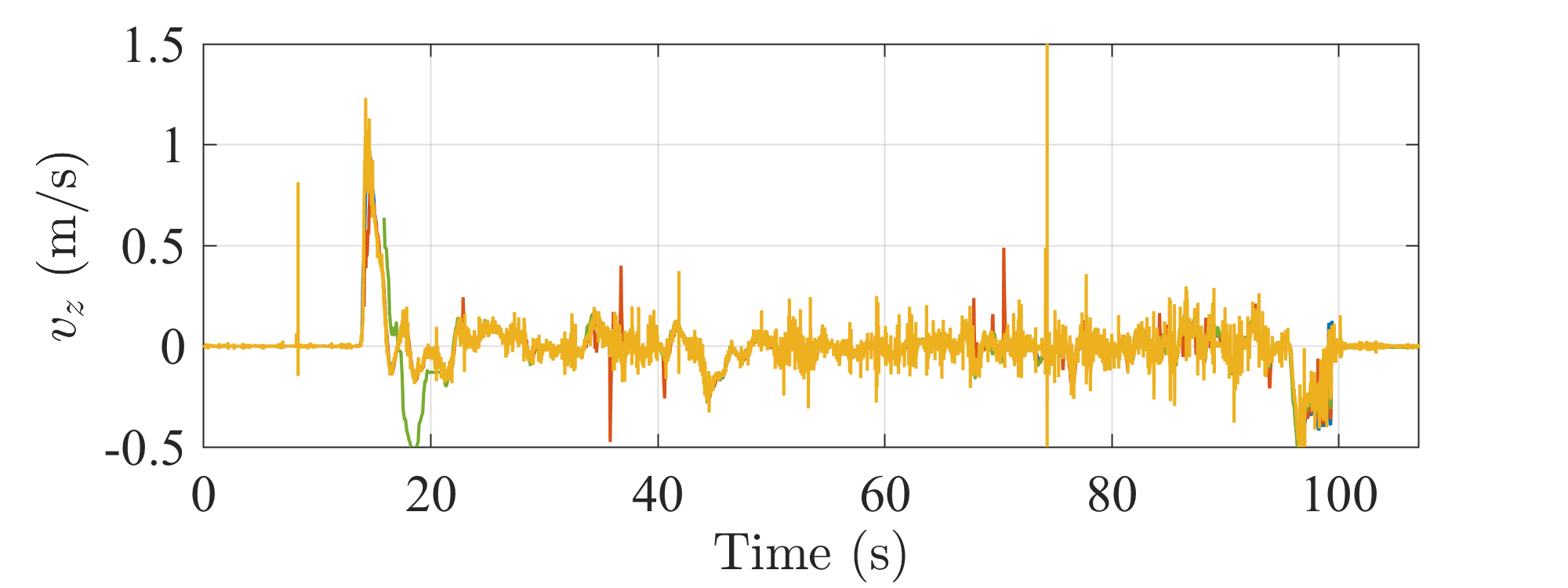} \\
	\caption{Comparison of velocities in MAV body frame in a real-world indoors dataset with motion capture ground truth.}	
	\label{fig:fig_real}
\end{figure}
			
\begin{figure}[h]
	\centering
		\includegraphics[width=0.9\linewidth]{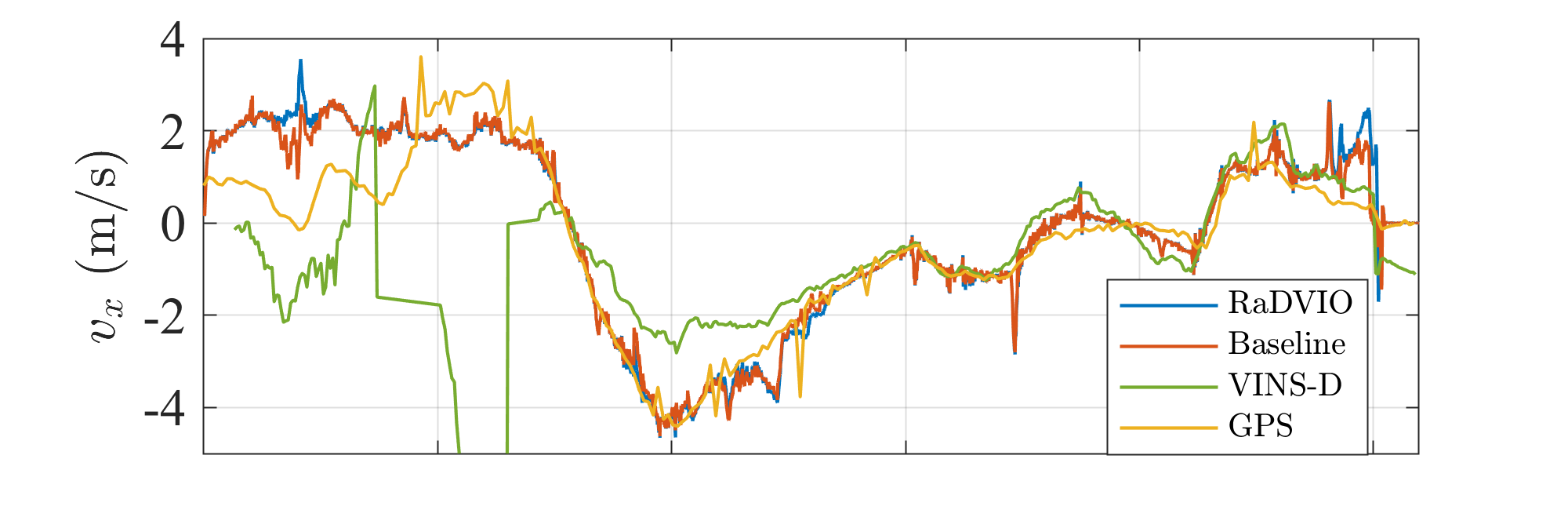}
		\includegraphics[width=0.9\linewidth]{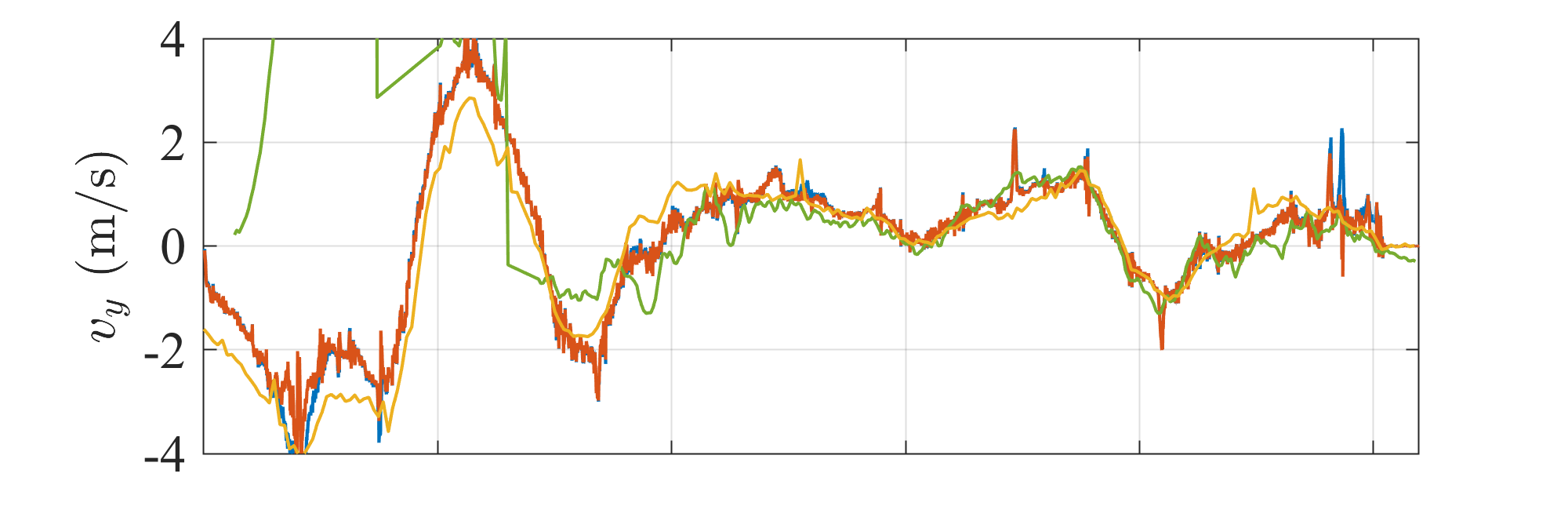}
		\includegraphics[width=0.9\linewidth]{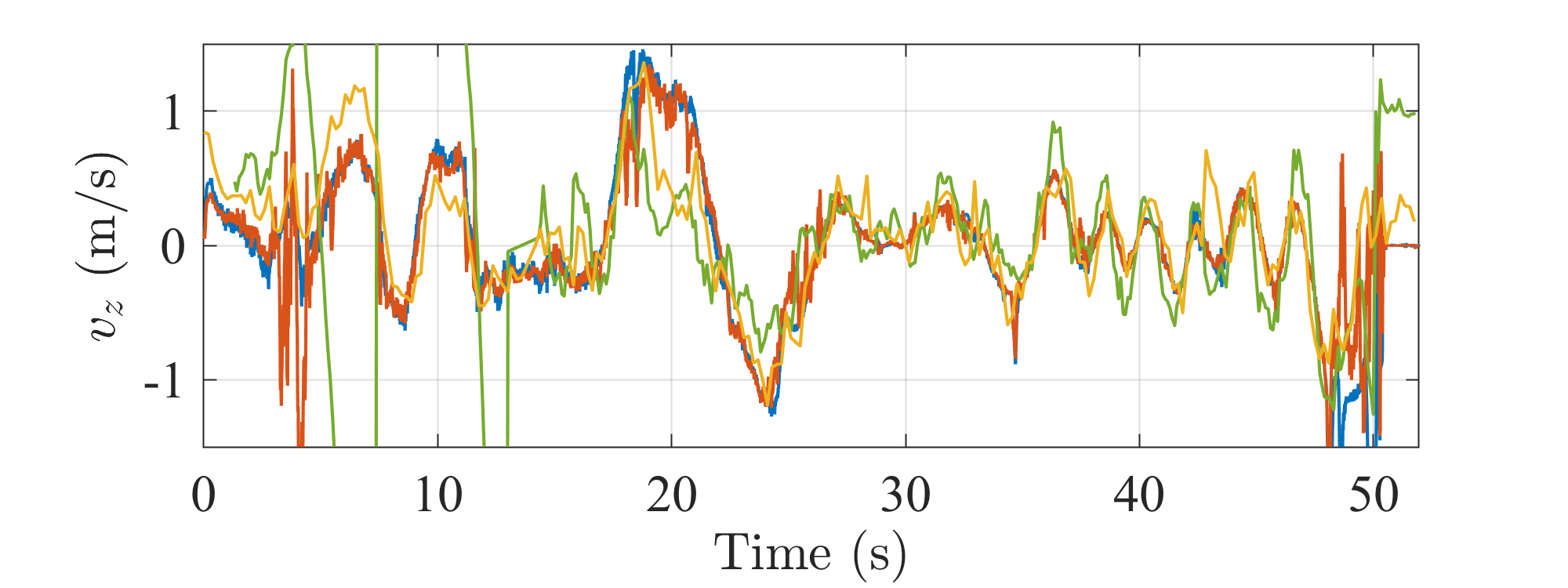}
	\caption{Comparison of velocities in MAV body frame for an outdoors trajectory (spans a $25m \times 45m$ region in a playground).}
	\label{fig:gps}
\end{figure}

The experiment results show that the proposed method is able to work well in real world even in the presence of sensor synchronisation issues and large noises in the IMU signal. 
																										
\subsection{Timing Benchmarks}
We evaluated the tracking framerate of RaD-VIO on a desktop PC with an Intel i7-6700K CPU and also onboard on an NVIDIA TX2. The frame rate of the tracker is on average 150Hz over all of our datasets on a PC and 55 Hz onboard, see Table~\ref{tab:hz_analysis}. Data sets with less features and large inter frame displacements result in lower frame rates due to longer time required for convergence.
\begin{table}[h]
	\captionsetup[table]{position=bottom}
	\centering
		\begin{tabular}{c|cccc}
			\toprule
			Category                    & Mean & $\sigma$ & Min & Max \\
			\hline
			Desktop                     & 153  & 15.7     & 107 & 181 \\
			Onboard                     & 55   & 7.0      & 40  & 65  \\
			\bottomrule
		\end{tabular}
	\caption{Frame Rate Evaluation.}\label{tab:hz_analysis}
\end{table}

%% file: ieeeconf/conclusion.tex
\section{Conclusions And Future Work}
\begin{table}
	\captionsetup[table]{position=bottom}
	\centering
		\begin{tabular}{l|ccc}
			\toprule
			Conditions               & Baseline & RaD-VIO  & VINS-D \\
			\hline
			Ideal (p1)               & ++       & +++* & +++    \\
			Low texture (p2)         & +        & +++  & +++    \\
			Negligible texture (p3)  & failure	& +++  & +      \\
			Moving features (p4, m2) & -        & -    & ++     \\
			Extreme motion (p5)      & -        & +++  & +++    \\
			Low image Hz (p6)        & -        & +++  & ++     \\
			Slope (s1)               & +++      & +++  & +++    \\
			Medium clutter (m1)      & +        & ++   & +++    \\
			High clutter (c1)        & +        & +    & +++    \\
			\bottomrule
		\end{tabular}
	
		\caption{Qualitative performance comparison on simulation data. +++:Low or no tracking failures, ++:Occasional failures, +:Frequent failures, -:Works poorly}\label{tab:qual_analysis}
\end{table}

We present a framework to obtain 6 degree of freedom state estimation on MAVs using a downward camera, IMU, and a downward single beam rangefinder. The proposed approach first extracts the rotation and unscaled translation between two consecutive image frames based on a cost function that combines dense photometric homography based alignment and a rotation prior from the IMU, and then uses an EKF to perform sensor fusion and output a filtered metric linear velocity. 

Extensive experiments in a wide variety of scenarios in simulations and in real-world experiments demonstrate the accuracy and robustness of the tracker under extenuating circumstances. The performance exceeds the frame-to-frame tracking framework proposed in~\cite{grabe2012board, grabe2015nonlinear}, and is slightly better than a current state of the art monocular visual-inertial odometry algorithm. Baseline fails in high clutter, extreme motion, or low texture, VINS-D fails when it doesn't initialise and in low texture. RaD-VIO degrades when in high clutter, but crucially it is stable and never generates extremely diverged state estimates (as triangulation based optimisation methods are susceptible to), and can operate at a high frame rate. The ability to run on high frame rate image streams ensures that consecutive images have high overlap and helps mitigate the common issue of poor performance when close to the ground. A qualitative comparison of the performance on simulation data is shown in Table~\ref{tab:qual_analysis}.

To relax the planar assumption in the proposed method, we replaced the SSD error between pixel intensities with Huber and Tukey loss functions~\cite{huber1964robust, maronna2006robust}. The accuracy improvement in cluttered environment was minor and offset by the increase in computational costs. In future work we aim to address the weakness of the planar assumption through means of explicitly accounting for it within the formulation.

Additionally, in line with our introductory statements, we intend to couple the performance of RaD-VIO with a conventional forward facing sliding window VIO algorithm to develop a resilient robotic system that exploits the individual strengths of both odometry approaches.